\documentclass{article}

\PassOptionsToPackage{numbers, compress}{natbib}
\usepackage[final]{neurips_data_2022}
\usepackage{graphicx}
\usepackage{times}
\usepackage{comment}
\usepackage{amsmath,amssymb,amsfonts,mathabx,mathbbol}
\usepackage{acronym}
\usepackage{enumitem}
\usepackage[pagebackref,colorlinks,breaklinks,citecolor=gray]{hyperref}
\usepackage{url}
\usepackage{balance}
\usepackage{xspace}
\usepackage{setspace}
\usepackage[skip=3pt,font=small]{subcaption}
\usepackage[skip=3pt,font=small]{caption}
\usepackage{caption}
\usepackage[misc]{ifsym}
\usepackage[dvipsnames]{xcolor}
\usepackage[capitalise]{cleveref}
\usepackage{tabularx}
\usepackage{multirow,multirow,array,makecell}
\usepackage{overpic}
\usepackage{wrapfig}
\usepackage{pifont}
\usepackage{booktabs,tabularx,colortbl}       % professional-quality tables
\usepackage[utf8]{inputenc} % allow utf-8 input
\usepackage[T1]{fontenc}    % use 8-bit T1 fonts
\usepackage{nicefrac}       % compact symbols for 1/2, etc.
\usepackage{microtype}      % microtypography
\usepackage[noend, ruled,vlined]{algorithm2e}
\usepackage{algorithmic}
\usepackage{rotating}
\usepackage{adjustbox}
\usepackage{hvfloat}

\makeatletter
\DeclareRobustCommand\onedot{\futurelet\@let@token\@onedot}
\def\@onedot{\ifx\@let@token.\else.\null\fi\xspace}
\def\eg{\emph{e.g}\onedot} 

\def\ie{\emph{i.e}\onedot}

\def\etc{\emph{etc}\onedot}

\makeatother

\newcommand{\dec}[1]{\textcolor{blue}{#1}}
\newcommand{\inc}[1]{\textcolor{red}{#1}}

\newcommand{\cmark}{\ding{51}}
\newcommand{\xmark}{\ding{55}}

\makeatletter
\renewcommand{\paragraph}{%
  \@startsection{paragraph}{4}%
  {\z@}{0ex \@plus 0ex \@minus 0ex}{-1em}%
  {\hskip\parindent\normalfont\normalsize\bfseries}%
}
\makeatother

\crefname{algorithm}{Alg.}{Algs.}
\Crefname{algocf}{Algorithm}{Algorithms}
\crefname{section}{Sec.}{Secs.}
\Crefname{section}{Section}{Sections}
\crefname{table}{Tab.}{Tabs.}
\Crefname{table}{Table}{Tables}
\crefname{figure}{Fig.}{Fig.}
\Crefname{figure}{Figure}{Figure}

\graphicspath{{figure/}}

\acrodef{nlp}[NLP]{Natural Language Processing}
\acrodef{amt}[AMT]{Amazon Mechanical Turk}
\acrodef{ucla}[UCLA]{University of California, Los Angeles}
\acrodef{pku}[PKU]{Peking University}
\acrodef{bigai}[BIGAI]{Beijing Institute of General Artificial Intelligence}

\definecolor{scope}{RGB}{103,78,167}
\definecolor{semantic}{RGB}{230,145,56}
\definecolor{type}{RGB}{153,0,0}

\newcommand\supp{\textit{supplementary}\xspace}

\makeatletter
\renewcommand*{\@fnsymbol}[1]{\ensuremath{\ifcase#1\or \dagger\or *\or \ddagger\or
   \mathsection\or \mathparagraph\or \|\or **\or \dagger\dagger
   \or \ddagger\ddagger \else\@ctrerr\fi}}
\makeatother

\title{EgoTaskQA: Understanding Human Tasks in Egocentric Videos}

\author{%
    Baoxiong Jia$^{1, 2}$\thanks{Work done during internship at BIGAI.} \\
    \small\texttt{baoxiongjia@ucla.edu} \\
    \And
    Ting Lei$^{2,3\dag}$ \\
    \small\texttt{ting\_lei@pku.edu.cn} \\
    \And
    Song-Chun Zhu$^{2,3,4}$ \\
    \small\texttt{sczhu@bigai.ai} \\
    \And
    Siyuan Huang$^{2}$ \\
    \small\texttt{syhuang@bigai.ai} \\
    \\
    $^1$UCLA Center for Vision, Cognition, Learning, and Autonomy (VCLA)\\
    $^2$ Beijing Institute for General Artificial Intelligence (BIGAI)\\
    $^3$ Institute for Artificial Intelligence, Peking University\\
    $^4$ Department of Automation, Tsinghua University\\
    \url{https://sites.google.com/view/egotaskqa}
}

\begin{document}

\maketitle

\begin{abstract}
Understanding human tasks through video observations is an essential capability of intelligent agents. The challenges of such capability lie in the difficulty of generating a detailed understanding of situated actions, their effects on object states (\ie, state changes), and their causal dependencies. These challenges are further aggravated by the natural parallelism from multi-tasking and partial observations in multi-agent collaboration. Most prior works leverage action localization or future prediction as an \textit{indirect} metric for evaluating such task understanding from videos. To make a \textit{direct} evaluation, we introduce the EgoTaskQA benchmark that provides a single home for the crucial dimensions of task understanding through question-answering on real-world egocentric videos. We meticulously design questions that target the understanding of (1) action dependencies and effects, (2) intents and goals, and (3) agents' beliefs about others. These questions are divided into four types, including descriptive (what status?), predictive (what will?), explanatory (what caused?), and counterfactual (what if?) to provide diagnostic analyses on \textit{spatial, temporal, and causal} understandings of goal-oriented tasks. We evaluate state-of-the-art video reasoning models on our benchmark and show their significant gaps between humans in understanding complex goal-oriented egocentric videos. We hope this effort will drive the vision community to move onward with goal-oriented video understanding and reasoning.

\end{abstract}

\section{Introduction}
\label{sec:intro}

The study of human motion perception has suggested that humans perceive motion as goal-directed behaviors rather than plain pattern movements~\cite{baldwin2001infants,gergely2002rational,woodward1998infants}. Developmental psychologists~\cite{csibra2007obsessed} categorized such an ability into two distinct mechanisms: (1) action-effect associations that the desired effects activate the corresponding action; and (2) simulative procedures, which argues that goal attribution comes from planning under the rational action principle in others' shoes. Both mechanisms require detailed knowledge of \textbf{action dependencies and effects}, agent's \textbf{intents and goals} and \textbf{beliefs about other agents}. With such knowledge playing crucial roles in human cognitive development, learning them from visual observation is pivotal for building more intelligent agents.

% (2) teleological reasoning, where predictions are governed by the rational action principle given situational constraints of the action being performed; and (3) simulation procedures, by simulatively generating the mental states the observers would posess were they in the other's shoes. These mechanisms further lead to three critical aspects of goal-oriented activity understanding, namely, \textbf{action-effect} learning, \textbf{situated reasoning} and multi-agent \textbf{belief modeling}. With all three mechanism contributing equally in human cognitive systems, such aspects are largely missing in the computational literature of egocentric video understanding or, more broadly, the field of computer vision.

Taking a closer look at how humans learn from interacting with the world, we locate objects, change their positions and manipulate them in various ways, all presumably under visual control from an egocentric perspective~\cite{land1999roles}. This unique first-person experience provides essential visual cues for human attention and goal-oriented task understanding. Moreover, egocentric perception naturally reflects how humans reason and perform in a partially observable environment, making it the most available learning source for learning actions, tasks~\cite{rajeswaran2022r3m}, and belief modeling~\cite{fan2021learning}.
% More importantly, the egocentric view reflects how human perceive and act when interacting with the world and collaborating with other agents in partially observable scenarios~\cite{land1999roles}. 
The past few years have witnessed significant progress in egocentric video understanding, especially action recognition and future anticipation~\cite{pirsiavash2012detecting,sigurdsson2018charades,li2018eye,jia2020lemma,damen2022rescaling,grauman2022ego4d}. However, these two tasks merely cover the tip of the iceberg, considering how humans learn from visual observations to obtain knowledge for more profound tasks such as learning world models, planning for desired goals, and building beliefs about others. With their essential roles in human cognitive development, we urge the need for a benchmark that addresses these missing dimensions in egocentric activity understanding.
\begin{figure}[t!]
    \centering
    \includegraphics[width=0.99\linewidth]{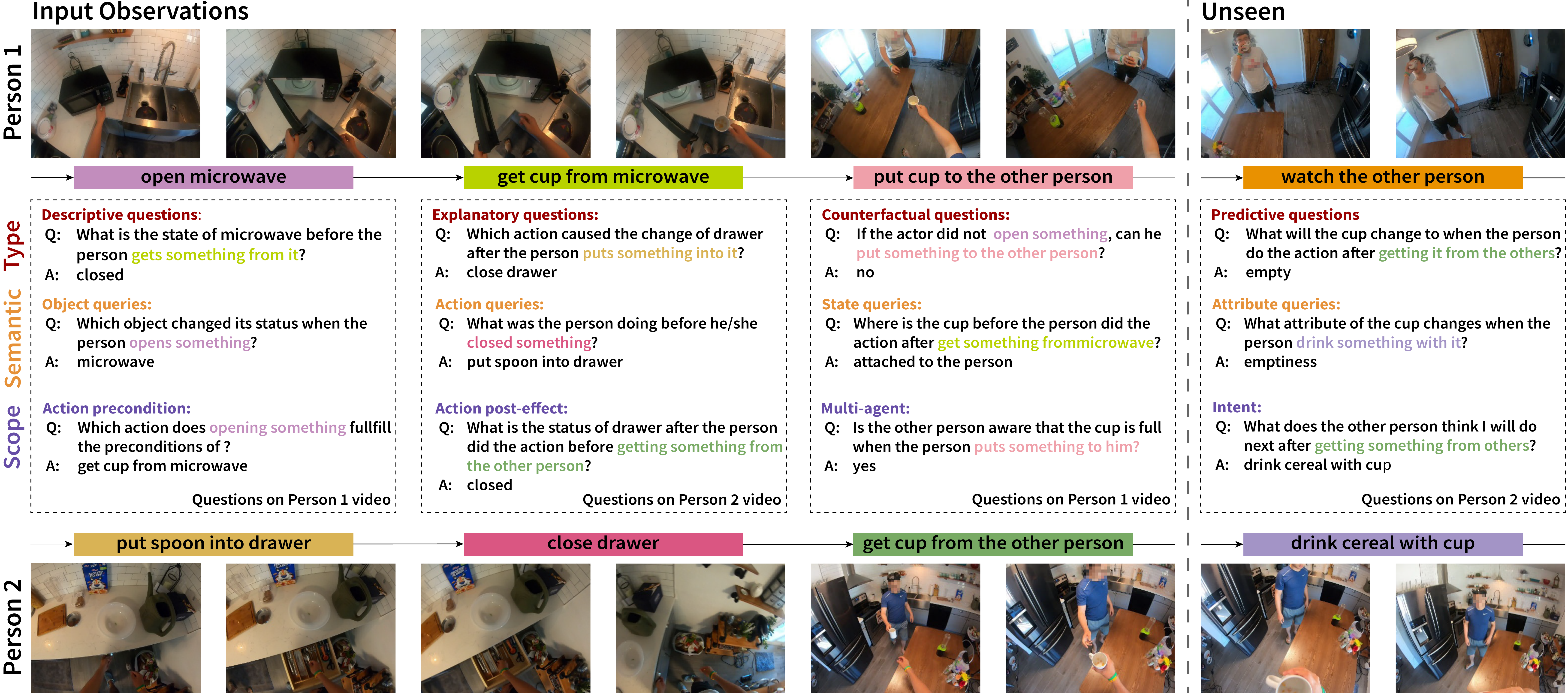}
    \caption{An overview of EgoTaskQA. We show an illustrative scenario where two subjects collaborate to make and drink cereal. Based on egocentric observations, we generate questions on these seen or unseen video intervals with different {\textcolor{type}{question types}}, {\textcolor{semantic}{targeting semantics}} and {\textcolor{scope}{question scopes}}. Note that we use both direct (\eg, when the person open something...) and indirect (\eg, the action before getting something...) references on actions and objects, where the same color indicates the same referred actions (best viewed in color).}
    \label{fig:overview}
\end{figure}
%Designing a benchmark that can properly evaluate action dependencies, effects and beliefs is inherently challenging due to the limitation of both data and annotations. Among existing attempts, the LEMMA dataset~\cite{jia2020lemma} shared similar insights on incorporating knowledge of both goal-oriented and multi-agent collaborative activities with fine-grained action and task annotation. Inspired by LEMMA and existing video reasoning benchmarks~\cite{yi2020clevrer,grunde-mcLaughlin2021agqa,wu2021star}, we propose a challenging egocentric, goal-oriented, video question-answering benchmark EgoTaskQA.

Hence, we present EgoTaskQA, a challenging egocentric, goal-oriented video question-answering benchmark based on the LEMMA dataset~\cite{jia2020lemma}. The LEMMA dataset collects egocentric videos in goal-oriented and multi-agent collaborative activities with fine-grained action and task annotation. By extending the LEMMA dataset with annotations consisting of object status, human-object and multi-agent relationships, and causal dependency structures between actions, we design questions that target three specific scopes: (1) actions with world state transitions and their dependencies, (2) agents' intents and goals in task execution, and (3) agents' belief about others in collaboration to provide an in-depth evaluation metric for task understanding. These questions are procedurally generated within four types: \textbf{descriptive}, \textbf{predictive}, \textbf{explanatory}, and \textbf{counterfactual}, to systematically test models' capabilities over \textbf{spatial}, \textbf{temporal}, and \textbf{causal} domains of goal-oriented task understanding. To avoid spurious correlations in questions, we include both direct and indirect references to actions and objects. We further balance the answer distribution by the reasoning type of questions and carefully design benchmarking train/test splits to provide a systematic test on goal-oriented reasoning and indirect reference understanding; see~\cref{fig:overview} for an example and more details in~\cref{sec:benchmark}.

As shown in~\cref{tab:dataset_comparison}, EgoTaskQA complements existing video reasoning benchmarks on various dimensions. With models exhibiting large performance gaps compared with humans, we devise diagnostic experiments to reveal both the easy and challenging spots in our benchmark. We hope such designs and analyses will foster new insights into goal-oriented activity understanding.

\paragraph{Contributions} In summary, our main contributions are three-fold:
\begin{itemize}[leftmargin=*,noitemsep,nolistsep]
\item We extend the LEMMA dataset with annotations of object status, human-object and multi-agent relationships to facilitate egocentric activity understanding. We further generate causal dependency structures between actions to provide ground truth for procedural task understanding.
\item We construct a balanced video question-answering benchmark, EgoTaskQA, to measure models' capability in understanding action dependencies and effects, intents and goals, as well as beliefs in multi-agent scenarios. We procedurally generate four challenging types of questions (descriptive, predictive, explanatory, and counterfactual) with both direct and indirect references for our benchmark and potential research on video-grounded compositional reasoning.
\item We devise challenging benchmarking splits over EgoTaskQA to provide a systematic evaluation of goal-oriented reasoning and indirect reference understanding. We experiment with various state-of-the-art video reasoning models, show their performance gap compared with humans, and Acknowledgementanalyze their strengths and weaknesses to promote future research on goal-oriented task understanding.
\end{itemize} 

\section{Related Work}\label{sec:related_work}

\paragraph{Action as Inverse Planning} Action understanding has been seen as an inverse planning problem on agents' mental states~\cite{baker2009action, shum2019theory}. Early studies formulate it as reasoning on the first-order logic formulae that describes actions' preconditions and post-effects~\cite{mccarthy1963situations,reiter1991frame}. This symbolic formalism is later paired with domain-specific language and algorithms to become mainstays in robotics planning~\cite{fikes1971strips,mcdermott1998pddl}. In computer vision, similar attempts have been made to link visual observations with world states and actions~\cite{duan2012discovering, isola2015discovering, nagarajan2018attributes}. Various methods treated actions as transformations on images to solve action-state recognition~\cite{fathi2013modeling, fire2015learning, wang2016actions, alayrac2017joint, liu2017jointly} and video prediction~\cite{oh2015action, vondrick2016generating, ha2018recurrent}. With the emerging interest in language-grounded understanding,~\citet{zellers2021piglet} proposed PIGLeT to study the binding between images, world states, and action descriptions.~\citet{padmakumar2022teach} further studies the problem of language understanding and task execution by designing an intelligent embodied agent that can chat during task execution. However, these works are mostly limited to atomic actions, missing the important action dependency in task execution. To tackle this  problem, instructional videos~\cite{kuehne2014language,alayrac2016unsupervised,tang2019coin,miech2019howto100m} are studied with its goal-oriented multi-step activities. In these videos, external knowledge~\cite{paulius2016functional, koupaee2018wikihow} can be used as guidance for advanced tasks like temporal dynamics learning~\cite{epstein2021learning} and visually grounded planning~\cite{chang2020procedure, sun2022plate}. Unfortunately, these videos highlight the instructions and include no task-level noise, which is much simpler than the partially observable, highly paralleled, multi-agent environment that humans learn from and as presented in our benchmark. These complexities make the goal-oriented action understanding a challenging task remaining to be solved.

\begin{table}[t!]
    \centering
    \caption{A comparison between EgoTaskQA and existing video question-answering benchmarks. We use ``world'' for world model-related information, including action preconditions, post-effects, and dependencies. We use FPV as short for egocentric and TPV for third-person-view videos. We use MC as short for multiple-choice question-answering, and OP for open-answer question-answering.}
    \resizebox{\linewidth}{!}{%
        \begin{tabular}{cccccccccccc}
        \toprule
        \multirow{2}[2]{*}{Dataset} & \multicolumn{2}{c}{Video} & \multicolumn{3}{c}{Question Scope} & \multicolumn{4}{c}{Question type} & \multirow{2}[2]{*}{\shortstack[c]{Answer \\ Type}} & \multirow{2}[2]{*}{\# questions} \\ 
        \cmidrule(lr){2-3}\cmidrule(lr){4-6}\cmidrule(lr){7-10}
         & View & Real-world & World & Intents \& Goals & Multi-agent & Descriptive & Predictive & Explanatory & Counterfactual &  & \\
        \midrule
        MarioQA~\cite{mun2017mario} & TPV & \xmark & \cmark & \xmark & \xmark & \cmark & \xmark & \cmark & \xmark  & OP & 188K \\
        Pororo-QA~\cite{kim2017deepstory} & TPV & \xmark & \cmark & \xmark & \xmark & \cmark & \xmark & \cmark & \xmark & MC & 9K \\
        CLEVRER~\cite{yi2020clevrer} & TPV & \xmark & \xmark & \xmark & \xmark & \cmark & \cmark & \cmark & \cmark & OP+MC & 282K \\
        Env-QA~\cite{gao2021envqa} & FPV & \xmark & \cmark & \xmark & \xmark & \cmark & \xmark & \xmark & \xmark & OP & 85K \\
        MovieQA~\cite{tapaswi2016movieqa} & TPV & \cmark & \xmark & \xmark & \xmark & \cmark & \xmark & \cmark & \xmark & MC & 14K \\
        Social-IQ~\cite{zadeh2019socialiq}  & TPV & \cmark & \xmark & \cmark & \cmark & \cmark & \xmark & \cmark & \xmark & MC & 7.5K \\
        TVQA~\cite{lei2018tvqa} & TPV & \cmark & \xmark & \xmark & \xmark & \cmark & \xmark & \cmark & \xmark & MC & 152.5K \\
        TVQA+~\cite{lei2020tvqa+} & TPV & \cmark & \xmark & \xmark & \xmark & \cmark & \xmark & \cmark & \xmark & MC & 29.4K \\
        MSVD-QA~\cite{xu2017video} & TPV & \cmark  & \xmark & \xmark & \xmark & \cmark & \xmark & \xmark & \xmark & OP & 50.5K \\
        MSRVTT-QA~\cite{xu2017video} & TPV & \cmark & \xmark & \xmark & \xmark & \cmark & \xmark & \xmark & \xmark & OP & 243K \\
        Video-QA~\cite{zeng2017leveraging} & TPV & \cmark & \xmark & \xmark & \xmark & \cmark & \xmark & \xmark & \xmark & OP & 175K \\
        ActivityNet-QA~\cite{yu2019activitynet-qa} & TPV & \cmark & \xmark & \xmark & \xmark & \cmark & \xmark & \xmark & \xmark & OP & 58K\\
        TGIF-QA~\cite{jang2017tgif-qa} & TPV & \cmark & \xmark & \xmark & \xmark & \cmark & \xmark & \xmark & \xmark & MC & 165.2K \\
        How2QA~\cite{li2020hero} & TPV & \cmark & \xmark & \xmark & \xmark & \cmark & \xmark & \xmark & \xmark & MC & 44K \\
        HowToVQA69M~\cite{yang2021just} & TPV & \cmark & \xmark & \xmark & \xmark & \cmark & \xmark & \xmark & \xmark & OP & 69M\\
        AGQA~\cite{grunde-mcLaughlin2021agqa} & TPV & \cmark & \xmark & \xmark & \xmark & \cmark & \xmark & \xmark & \xmark & OP & 3.6M\\
        NExT-QA~\cite{xiao2021next} & TPV & \cmark & \xmark & \cmark & \xmark & \cmark & \cmark & \cmark & \xmark &  OP+MC & 52K\\
        STAR~\cite{wu2021star} & TPV & \cmark & \cmark & \xmark & \xmark & \cmark & \cmark & \xmark & \xmark &  MC & 60K\\
        EgoVQA~\cite{fan2019egovqa} & FPV & \cmark & \xmark & \xmark & \cmark & \cmark & \xmark & \xmark & \xmark & OP+MC & 520\\
        % Ego4d~\cite{grauman2022ego4d} & FPV & \cmark & \xmark & \xmark & \cmark & \xmark & \cmark & \xmark & \xmark & Query & TBD \\
        \midrule
        EgoTaskQA (Ours) & FPV & \cmark & \cmark & \cmark & \cmark  & \cmark & \cmark & \cmark & \cmark & OP & 40K\\
        \bottomrule
        \end{tabular}
    }
    \label{tab:dataset_comparison}
\end{table}

\paragraph{Egocentric Vision} Egocentric vision offers a unique perspective for actively engaging with the world. Aside from traditional video understanding tasks like video summarization~\cite{lee2012discovering, lu2013story}, activity recognition~\cite{feichtenhofer2019slowfast,wu2022memvit, girdhar2022omnivore} and future anticipation~\cite{nagarajan2020ego,furnari2020rolling,qi2018generalized,qi2020generalized,girdhar2021anticipative}, egocentric videos provide fine-grained information for tasks like human-object interaction understanding~\cite{nagarajan2019grounded,damen2016you,cai2016understanding,bambach2015lending,qi2018learning,garcia2018first,ma2016going} and gaze/attention prediction~\cite{wei2018and,li2018eye}. With its natural reflectance of partial observability, egocentric videos are also used for social understanding tasks such as joint attention modeling~\cite{fathi2012social, soo2015social}, perspective taking~\cite{yagi2018future, ng2020you2me} and communicative modeling~\cite{northcutt2020egocom, fan2021learning}. However, with various egocentric datasets curated over the last decade~\cite{pirsiavash2012detecting,lee2012discovering,sigurdsson2018charades}, data and detailed annotations for human tasks are still largely missing. Large-scale daily lifelog datasets like EPIC-KITCHENS~\cite{damen2022rescaling} and Ego4D~\cite{grauman2022ego4d} cover certain aspects of action-dependencies, effects, and social scenarios in their recordings, but are unsuitable for detailed annotation due to their size. The other stream of datasets collects activities by providing coarse task instructions to both single actor~\cite{rai2021home} and multiple agent collaborations~\cite{jia2020lemma}. They annotate tasks and compositional actions to reveal agents' execution and collaboration process for multi-step goal-directed tasks. Despite all the preferred characteristics of these goal-oriented activity videos, none of them successfully addressed action-dependencies and effects, nor multi-agent belief modeling.

\paragraph{Video Question-Answering Benchmarks} Visual question-answering can be designed to evaluate a wide spectrum of model capabilities, spanning from visual concept recognition and spatial relationship reasoning~\cite{tu2014joint,antol2015vqa, johnson2017clevr,hudson2019gqa}, abstract reasoning~\cite{barrett2018measuring,zhang2019raven,nie2020bongard,zhang2021abstract,zhang2021acre,zhang2022learning}, to common sense reasoning~\cite{park2020visualcomet, zellers2019recognition}. In the temporal domain, synthetic environments are used for questions that involve simple action-effect reasoning~\cite{mun2017mario, kim2017deepstory}. Crowdsourced videos~\cite{jang2017tgif-qa,yu2019activitynet-qa,lei2018tvqa,yang2021just} are used for collecting questions on basic spatial-temporal reasoning capabilities like event counting~\cite{jang2017tgif-qa}, grounding~\cite{lei2020tvqa+}, and episodic memory~\cite{grauman2022ego4d}. Recent advances in video question-answering aim for more profound reasoning capabilities. \citet{gao2021envqa} leverages an indoor synthetic environment to generate questions on spatial relationships and simple action-effect reasoning from an egocentric perspective. \citet{xiao2021next} designs NExT-QA containing questions about knowledge of the past, present, and future on both temporal and causal domains. \citet{grunde-mcLaughlin2021agqa} programmatically generates questions for compositional spatial-temporal reasoning and generalization. \citet{wu2021star} focus on short atomic action clips for situated reasoning. \citet{yi2020clevrer} generates synthetic videos for studying counterfactual predictions on collisions. \citet{zadeh2019socialiq} collects questions for social intelligence evaluation. Nevertheless, none of these benchmarks addressed the aforementioned critical dimensions of goal-oriented activity understanding from a real-world egocentric perspective.

\section{The EgoTaskQA Benchmark}\label{sec:benchmark}
The EgoTaskQA benchmark contains 40K balanced question-answer pairs selected from 368K programmatically generated questions generated over 2K egocentric videos. We target the crucial dimensions for understanding goal-oriented human tasks, including action effects and dependencies, intent and goals, and multi-agent belief modeling. We further evaluate models' capabilities to describe, explain, anticipate, and make counterfactual predictions about goal-oriented events.
A detailed comparison between EgoTaskQA and existing benchmarks is shown in~\cref{tab:dataset_comparison}. 

\subsection{Data Collection}\label{sec:data:annotation}
We select egocentric videos from the LEMMA dataset~\cite{jia2020lemma} as base video sources. Compared to similar egocentric datasets, human activities in LEMMA are highly goal-oriented and multi-tasked. These activities contain rich human-object interactions and action dependencies in both single-agent and two-agent collaboration scenarios. We take advantage of these desired characteristics and augment LEMMA with ground truths of object states, relationships, and agents' beliefs about others. More specifically, we augment LEMMA on the following aspects:

\begin{figure}[t!]
    \centering
    \includegraphics[width=\linewidth]{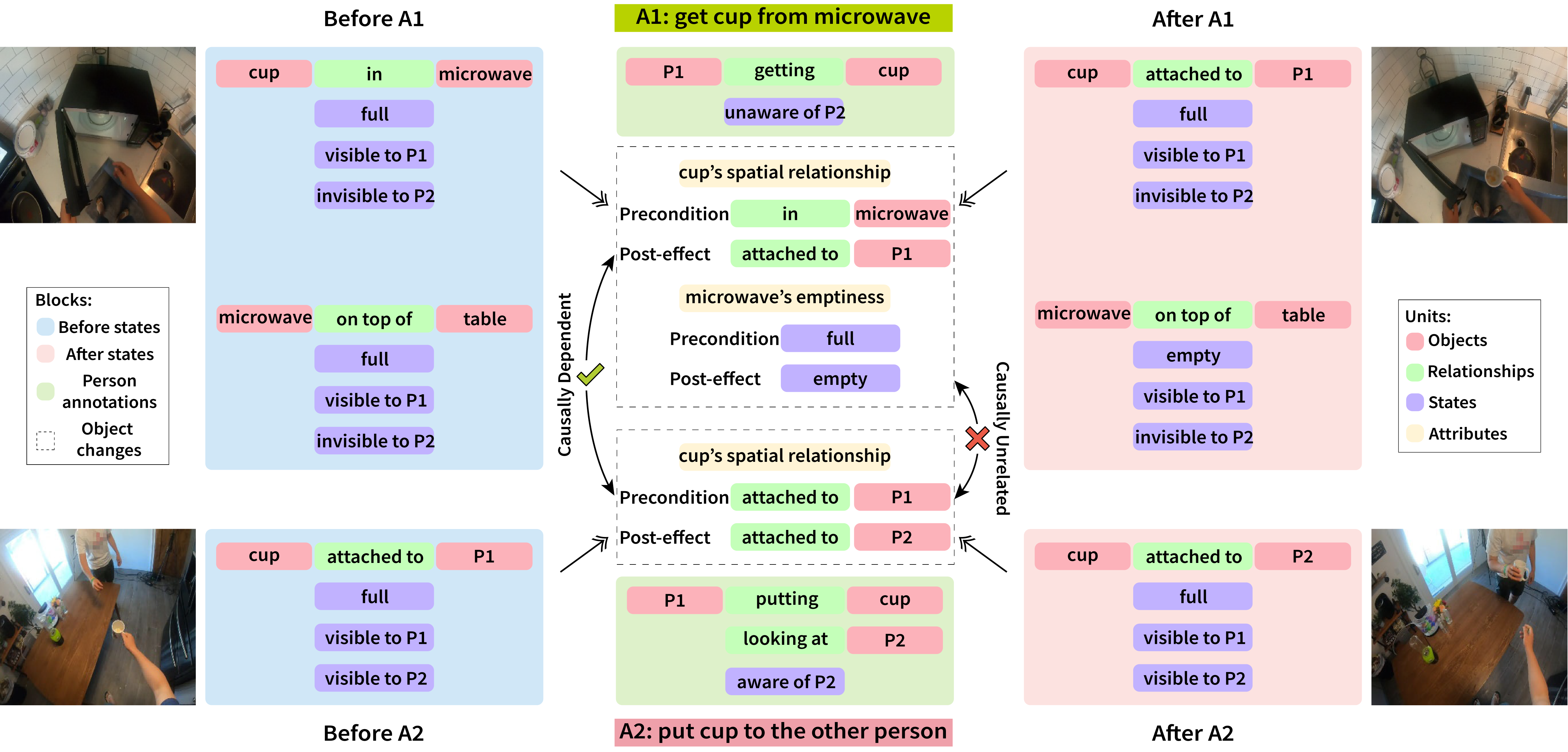}
    \caption{We use two actions A1:``get cup from microwave'' and A2:``put cup to the other person'' from Person 1's video in~\cref{fig:overview} as an example to visualize annotations in EgoTaskQA. We annotate states and relationships for objects changed by actions as well as human-object and multi-agent relationships. After obtaining the ``before'' and ``after'' annotations, we examine which attributes of objects were changed by the action and what are the preconditions and post-effects. We determine the causal dependency between actions by checking if there exists an object that the post-effect of one action over this object fulfills the preconditions of another. In this case, the state change of ``cup'' determines that A1 and A2 are causally dependent (best viewed in color).}
    \label{fig:data_collection_stats}
\end{figure}

\paragraph{World States} We focus on world states consisting of object states, object-object relationships, and human-object relationships. First, we build the vocabulary of relationships and state attributes from activity knowledge defined in previous works~\cite{paulius2016functional, ji2020action}. We manually filter irrelevant relationships and attributes by removing dataset-specific (\eg, under the car) and detailed numerical (\eg, cut in three) relationships. Next, we gather similar relationships to obtain 48 relationships and 14 object attributes. This vocabulary covers spatial relationships (\eg, on top of), object affordances (\eg, openable), and time-varying attributes (\eg, shape). We build on top of action annotations from LEMMA and use~\ac{amt} to annotate this information before and after the changing action for all time-varying objects. With these annotations, we reconstruct the transition chain for each interacted object and obtain their temporal status. We provide the complete list of relationships and object attributes in the \supp.

\paragraph{Multi-agent Relationships} To capture how two agents (actor and helper) collaborate over the same task, we annotate basic information about objects' visibility and the actor's awareness of the helper. For each object that the actor operates on, we annotate its visibility to the helper by providing synchronized videos from both agents' views to~\ac{amt} workers. For the actor's awareness of others, we instruct \ac{amt} workers to first go through the egocentric view video of both agents to get familiar with actions performed by the actor and the helper. Next, we ask \ac{amt} workers to replay the video of the actor and annotate, during each action segment, whether the actor can see the helper or whether the actor is aware of the helper's action if the helper is not in sight. As this annotation is usually subjective, we take the majority vote of three workers as ground truth.

\paragraph{Causal Trace} Based on the annotated transition chain of objects, we generate causal traces for each action with rules. By checking whether the post-effect of one action fulfills the preconditions of another, we define the causal relationship between two actions into unrelated, related, and causally dependent; see~\cref{fig:data_collection_stats} for an illustration and refer to \supp for detailed explanations. Given a video, we run this dependency check for each pair of actions. Next, we generate a video-level dependency tree by recursively checking sequential depending relationships and use it as the ground truth dependency structure for subsequent explanatory and counterfactual question generation.

In total, we augment LEMMA with 30K annotated before states, after states, and person annotation blocks as shown in~\cref{fig:data_collection_stats}. We then segment the videos in LEMMA into clips with lengths of around 25 seconds for question generation. This design helps generate interesting clips with partially observed environmental constraints (\eg, the cup is already washed when the person pours juice), and visual hints for future actions (\eg, cutting watermelon into dice instead of pieces for making juice rather than eating it directly). Meanwhile, we keep our videos reasonably long, with an average of 5 actions per clip to cover sufficient information for action dependency inference and future prediction. We provide more details about data collection and annotation statistics in \supp.

\subsection{Question-Answer Generation}\label{sec:qa:generation}
We use machine-generated questions to evaluate models' task understanding capabilities. We focus on the transition chain of each interacted object, especially what actions caused changes on objects and how these changes contribute together to a multi-step task; see examples in~\cref{fig:overview}. 

\paragraph{Question Design} We design questions that pinpoint scopes, including (1) action preconditions, post-effects, and their dependencies, (2) agents' intents and goals, and (3) agents' beliefs about others. Similar to~\cite{yi2020clevrer}, we categorize our questions over these three scopes into four types to systematically test models' capabilities over spatial, temporal, and causal domains of task understanding:
\begin{itemize}[leftmargin=*,noitemsep,nolistsep]
\item{\textbf{Descriptive}} questions evaluate the understanding of detailed spatial-temporal information. We provide spatial-temporal references in the questions to identify a unique interval for answering queries on objects and actions. These properties include object states and changes, relationships, human actions, and multi-agent-related information. We generate this type of question by randomly sampling an interval in the video clip and then gathering all related annotations for question generation. Answers in this category are generated based on the interval annotation and contain both open-ended queries and statement verifications.
\item{\textbf{Predictive}} questions aim at understanding intents and task planning. Given a video clip, we ask about possible future object states and actions for both the actor and the helper. These predictions include both direct predictions on actions and objects, as well as more challenging task-dependent predictions such as the executability of actions and the desired states of objects. Questions and answers for predictive questions are generated by gathering the future action/object annotations in a fixed window size after the truncated interval (\ie unseen future video) in the long original video. Answers in this category are open-ended action, object, and state queries.
\item{\textbf{Counterfactual}} questions aim at understanding action preconditions and post-effects. Based on the causal trace of actions, we generate counterfactual questions with hypothetical conditions that certain actions in the sequence were not executed. Under this condition, we query both the affected and unaffected actions about their executability and whether the corresponding changes of object states associated with these actions will occur. We generate counterfactual questions by adding or removing actions in the causal trace and adjusting the depending actions' executability recursively. Answers in this category contain action executability verifications and object state queries.
\item{\textbf{Explanatory}} questions evaluate the understanding of task-related object changes as well as action preconditions and post-effects. Given the object state annotations and the causal trace, we query the cause of state changes, the leading factor that satisfies the preconditions of specific actions, as well as why would the post-effect of certain actions affects other actions in the video clip.
% We design questions like "which action caused the cup's status to change to clean?", "How would wash change the status of cup?" and "Does cut watermelon fulfill the preconditions of eating watermelon?". 
We generate explanatory questions by querying both the annotations as well as the causal trace. Answers for explanatory questions contain both open-ended and verification queries.
\end{itemize}

\begin{figure}[t!]
    \includegraphics[width=\linewidth]{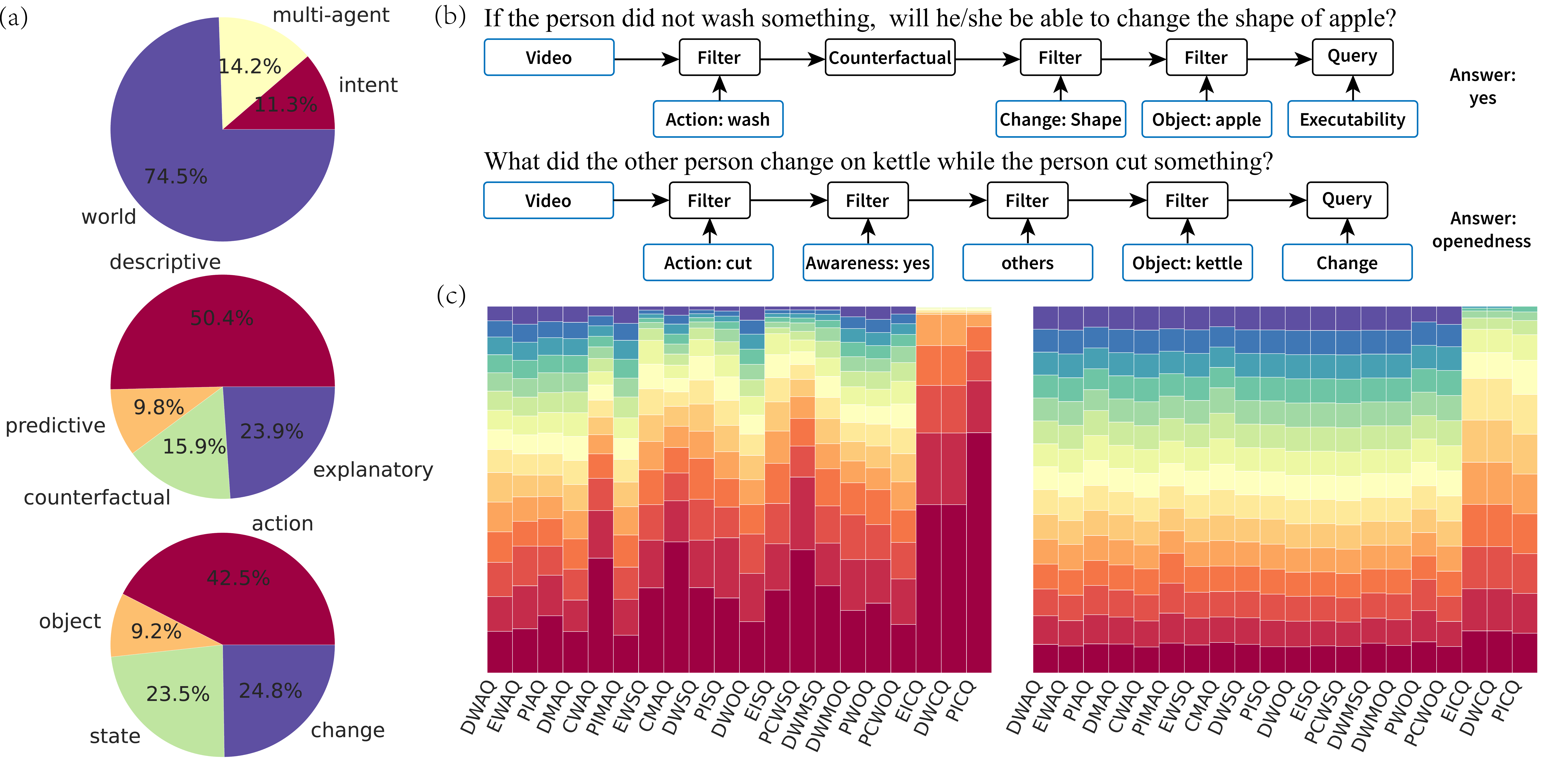}
    \caption{Generation and statistics of the question-answer pairs. (a) Question distribution according to question scope (top), question type (middle), and targeting semantics (bottom). (b) Examples of natural language questions and their corresponding executable programs. Operators and parameters of the program are represented by black and blue contour rectangles, respectively. (c) Answer distribution for the top 20 reasoning types and top 15 answers for each type before (left) and after (right) balancing, the reasoning types are abbreviated with the concatenation of their initial letters (\eg, DWAQ for descriptive, world, action, and query).}
    \label{fig:qa_generation_stats}
\end{figure}

\paragraph{Answer Generation} In EgoTaskQA, we consider both open-answer queries and binary statement verifications. To generate question-answer pairs for these questions, we design both text templates and the corresponding functional program templates as shown in~\cref{fig:qa_generation_stats} (b). Each program consists of a sequence of modules for querying the answer from the annotations and the causal trace. We exhaustively execute all possible program instantiations on videos to obtain answers by substituting arguments with instances in the available sample space. As all questions take action grounding as a prerequisite, we add indirect references (\eg, the first..., the action before...) to actions and objects when making substitutions to reflect this challenge; see details in \supp. After this initial process, we obtain 368K question-answer pairs over 2K videos as the full question set. 
% For open-answer problems, we focus on queries of actions, objects, states, and changing attributes.

% the actions from both the actor and the helper as well as objects and their state attributes for question sample space. We instantiate question templates with each combination of argument substitutions and execute the program on the provided annotations to obtain the ground truth answer. By exhaustively executing all program instantiations on the available videos, we obtian 380K question pairs for balancing.\bx{object semantic category, action, object, which attribute could change, attribute states}

\paragraph{Answer Distribution Balancing} We balance our answer distribution to avoid shortcuts from exploiting imbalances. Following the scheme introduced in~\cite{grunde-mcLaughlin2021agqa}, we tag each question template with its scope, type, and the targeting semantic category (\eg, actions, objects, states) and use the composition of all tags as the unique reasoning type for each question. We balance binary verification questions to have an equal proportion of each answer within each reasoning type. For open-answer questions, we use rejection sampling to ensure that the top 20\% frequent answers for each reasoning type do not appear as answers for more than 33.3\% of questions in the same type. After balancing by reasoning types, we proportionally sample questions to obtain a 40K diverse and balanced question set with a 1:2 ratio of binary and open-answer questions. We visualize the statistics of questions and answers and the effect of answer balancing in~\cref{fig:qa_generation_stats}. More details are provided in \supp.

\paragraph{Benchmark Splits}\label{para:split} We provide two benchmarking splits \textit{normal} and \textit{indirect} for video question-answering on EgoTaskQA. For the \textit{normal} split, we randomly sample questions according to their answer distribution and reasoning types to have a 3:1:1 split over training, validation, and test sets. The \textit{indirect} split is motivated by the fact that during task execution, actions, objects, and their changes are often strongly correlated. It leaves the chance for the model to perform well by simply over-fitting these strong correlations without thorough task understanding; see~\cref{sec:exp:text_modality} for a more in-depth discussion. We leverage the indirect references in our question to inspect the models' capability to use the learned knowledge for multi-step reasoning and generalize them to indirect references without over-fitting. More specifically, we filter questions without indirect references and simple indirect reference questions without multiple reasoning steps (\eg, what is the first action this person did? what did the person do before action ``putting something''?) from all question-answer pairs to form the training set, and split all indirect reference questions with multiple reasoning steps as validation and test sets. Under this setting, the \textit{indirect} split has a portion of 2:1:1 for training, validation, and test sets, respectively. We leave the remaining discussion of the \textit{indirect} split to~\cref{sec:exp:indirect}.

\section{Experiments}\label{sec:exp}
In this section, we evaluate and analyze the performance of video question-answering models on EgoTaskQA. We report how well models perform on different question scopes, types as well as targeting semantics on both \textit{normal} and \textit{indirect} splits. We also provide diagnostic experiments on the language modality to show the necessity of the \textit{indirect} split.
% We also show how text-based task knowledge could be helpful for answering these questions and report details of model performance on the more challenging indirect reference split. We treat the training of the EgoTaskQA question-answering as open answer answer classification problems for both query questions as well as verification questions. 

\paragraph{Baselines} In our experiments, we evaluate six state-of-the-art video question-answering models: VisualBERT~\cite{li2019visualbert}, PSAC~\cite{li2019beyond}, HME~\cite{fan2019heterogeneous}, HGA~\cite{jiang2020reasoning},
HCRN~\cite{le2020hierarchical}, and ClipBERT~\cite{lei2021less}. VisualBERT is a VL-BERT model designed for vision-language tasks. PSAC uses positional self-attention and co-attention network blocks to fuse visual and language features. HME uses external memory blocks for both visual inputs and questions on top of an LSTM-based encoder-decoder structure. HGA formulates video question-answering by constructing graphs for both videos and questions and aligning them. HCRN adopts a hierarchical framework by stacking relational modules over motion, question, and visual features. ClipBERT leverages sparsely sampled video clips and grid features~\cite{jiang2020defense} in a transformer architecture and achieves state-of-the-art results on video question-answering. We formulate question-answering in EgoTaskQA as a classification problem over all answer vocabulary and use models' accuracy as the evaluation metric under different settings. We provide details on model implementation, hyperparameter selection, and the training process in \supp.

\begin{table}[t!]
\caption{Model performance on the EgoTaskQA \textit{normal} split.}
\label{tab:direct_split}
\centering
\resizebox{\linewidth}{!}{%
\begin{tabular}{cccccccccc}
\toprule
 & Category & Most Likely & VisualBERT~\cite{li2019visualbert} & PSAC~\cite{li2019beyond} & HME~\cite{fan2019heterogeneous} & HGA~\cite{jiang2020reasoning} & HCRN~\cite{le2020hierarchical} & ClipBERT~\cite{lei2021less} & Human \\
\midrule
\multirow{3}{*}{\rotatebox[origin=c]{90}{Scope}} & world & 18.62 & 39.73 & 40.76 & 41.91 & 38.82 & \textbf{44.27} & 42.15 & 74\\
& intent & 2.54 & 44.51 & 46.19 & 48.92 & 42.12 & \textbf{49.77} & 40.94 & 82\\
& multi-agent & 10.92 & 26.29 & 30.59 & 27.98 & 23.43 & \textbf{31.36} & 27.63 & 76\\
\midrule
\multirow{4}{*}{\rotatebox[origin=c]{90}{Type}} & descriptive & 18.64 & 41.99 & 40.63 & 41.45 & 38.04 & \textbf{43.48} & 38.45 & 88\\
& predictive & 1.57 & 30.37 & 31.98 & 35.88 & 25.57 & \textbf{36.56} & 31.50 & 88\\
& counterfactual & 23.62 & 41.99 & 41.89 & 44.13 & 41.94 & \textbf{48.00} & 46.75 & 80\\
& explanatory & 7.97 & 37.42 & 37.99 & 38.85 & 35.97 & 40.60 & \textbf{42.39} & 74\\
\midrule
\multirow{4}{*}{\rotatebox[origin=c]{90}{Semantic}} & action & 10.05 & 15.02 & 14.75 & 14.99 & 15.08 & 14.92 & \textbf{22.91} & 70\\
& object & 2.07 & 23.26 & 36.53 & 36.05 & 19.09 & \textbf{45.31} & 21.80 & 82\\
& state & 6.05 & 59.20 & 61.89 & 63.44 & 55.65 & \textbf{68.28} & 54.36 & 80\\
& change & 41.97 & 68.27 & 65.05 & \textbf{68.87} & 68.38 & 67.38 & 66.58 & 82\\
\midrule
\multirow{3}{*}{\rotatebox[origin=c]{90}{Overall}} & open & 0.70 & 24.62 & 26.97 & 27.66 & 22.75 & \textbf{30.23} & 27.70 & 82\\
& binary & 50.46 & 68.08 & 65.95 & 68.6 & 68.53 & \textbf{69.42} & 67.52 & 76\\
& all & 15.4 & 37.93 & 38.90 & 40.16 & 36.77 & \textbf{42.20} & 39.87 & 80\\
\bottomrule
\end{tabular}
}
\end{table}

\subsection{Comparative Analysis}
We provide experimental results of baseline models on the EgoTaskQA \textit{normal} split in~\cref{tab:direct_split}. Model performances are evaluated on question scopes, types, targeting semantics, and overall answer categories. To quantify the naturalness and correctness of questions and answers in the EgoTaskQA benchmark, we provide human evaluation following the consistency check introduced in~\cite{hudson2019gqa, grunde-mcLaughlin2021agqa}. More specifically, we randomly sample 50 questions for each category and instruct \ac{amt} workers to evaluate the quality of the generated answer. Additionally, we compare all baseline models with a simple frequency-based baseline, namely "Most Likely" in~\cref{tab:direct_split}, where we select the most likely answer for each category to answer all questions in that category.  

As shown in~\cref{tab:direct_split}, the low performance of the most likely answer proves that our answer distribution is correctly balanced. For certain categories (\eg, change), the most likely answer has relatively high accuracy (41.97\%) as it covers both open-answer and binary questions. Next,
we observe relative low human performance in certain categories (\eg, action and explanatory). This indicates that identifying causal dependency between actions and conducting multi-step reasoning is not a trivial task for humans as also discovered in~\cite{grunde-mcLaughlin2021agqa}. However, we still observe a large gap between state-of-the-art models and human performance. Among all models, we find HME, HCRN, and ClipBERT to perform the best. This result is reasonable since they leverage different ways to provide better visual representations and interactions between video and language. Among all question scopes, we recognize a relatively low accuracy on multi-agent-related questions among all question scopes. It implies that understanding other agents' actions during task execution is still difficult without explicit modeling. It is significant in egocentric vision as a person's view changes dramatically, and only glances can be taken to acquire others' information. Meanwhile, we notice that these models perform relatively well for questions on states and changing attributes. We conjecture that this is attributed to the task knowledge embedded in textual descriptions of questions since actions, objects, and state changes are strongly correlated, as mentioned in~\cref{para:split}.

\subsection{The Effectiveness of Language}\label{sec:exp:text_modality}

\paragraph{Object information} We found the object information in the texts to be highly beneficial for question-answering on task-related knowledge during initial experiments. Compared to the original LEMMA action annotation (\eg, drinking [cereal] with [cup]), we use verbs to refer to actions in EgoTaskQA and obfuscate object information at different levels (\eg, drink something with cup, drink something with something) as similarly done in~\cite{grunde-mcLaughlin2021agqa,wu2021star}. While both types of action references localize to the same action interval, it contains different levels of knowledge in the language modality. Intuitively, the combination of action verbs (\eg, cut) and targeting objects (\eg, watermelon) provide object state information (\eg, diced) under certain scenarios. Therefore, we compare models' performance at different levels of object information obfuscation. As shown in~\cref{fig:obfuscation}, we recognize a significant performance gain for all models by gradually removing object information obfuscation in text, \ie, substituting ``something'' with the original object. This result supports the hypothesis that with fine-grained action annotations, we can learn task-related knowledge reasonably well by simply exploiting texts. It shares the same conclusion with recent works on leveraging text-based knowledge for helping instructional video understanding~\cite{lin2022learning}. To further investigate the effectiveness of the language modality, we conduct ablative experiments on the EgoTaskQA \textit{normal} split.

\paragraph{Language-Only} Language has been shown to provide knowledge that helps visual question-answering~\cite{goyal2017making}. To study the role of language in EgoTaskQA, we design a text-only setting for VisualBERT and HCRN, testing BERT~\cite{devlin2018bert} and HCRN without vision against their vision-language counterparts. As shown in~\cref{tab:without_vision}, the performance for most question categories dropped significantly. For the task of video question-answering, we should expect that dropping the vision branch will significantly affect the models' performance. As shown in~\cref{tab:without_vision}, we observe the general performance for the two models decreased as we expected. Among all categories, the models' performance for the objects decreased the most, which is consistent with the fact that the object queries highly depend on the situation provided in the videos (\eg, which object changed its status in the video?). However, we observe a slight performance gain on object state change questions. This further suggests that the knowledge of world state change, \ie which object attribute could change under actions, is embedded within question texts. Models could exploit question texts to learn simple associations between attribute types and action verbs (\eg, cleanliness and wash, emptiness and pour, shape and cut, \etc).

\begin{table}[t!]
\centering
\begin{minipage}{0.52\linewidth}
    \centering
    \includegraphics[width=\linewidth]{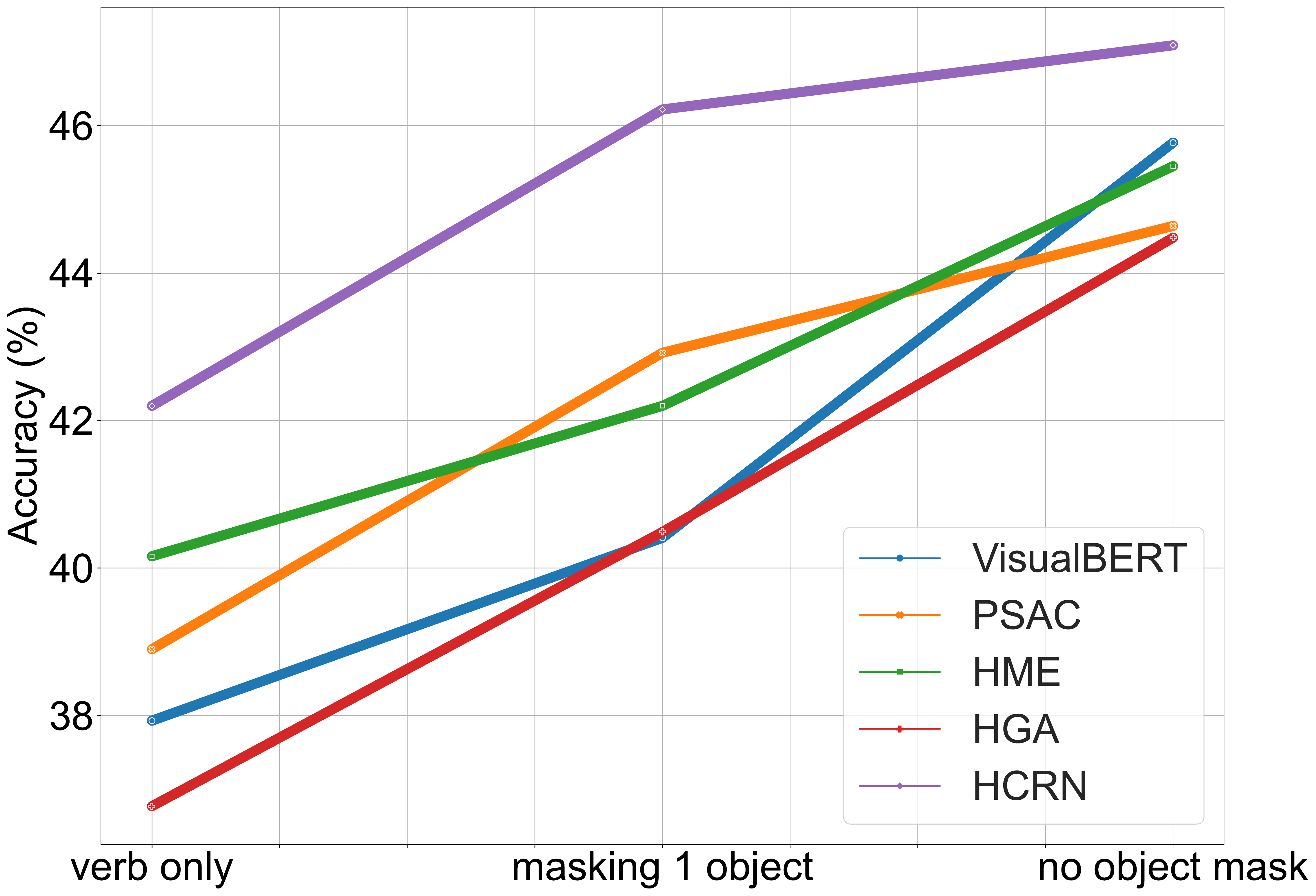}
    \captionof{figure}{Ablative study on model performance with different levels of object information obfuscation on the EgoTaskQA \textit{normal} split.}
    \label{fig:obfuscation}
\end{minipage}
\hfill
\begin{minipage}{0.46\linewidth}
\captionof{table}{Language-only question-answering results on the EgoTaskQA \textit{normal} split.}
\label{tab:without_vision}
\resizebox{\linewidth}{!}{
\begin{tabular}{ccccc}
\toprule
 \multirow{2}{*}{Category} & \multicolumn{2}{c}{BERT~\cite{devlin2018bert}} & \multicolumn{2}{c}{HCRN (w/o vision)}\\
 \cmidrule(lr){2-3}\cmidrule(lr){4-5}
 & Acc. & Change & Acc.& Change\\
 \midrule
world &  36.28 & \dec{-8.7\%} & 35.22 & \dec{-20.4\%}\\
intent & 35.02 & \dec{-21.3\%} & 34.93 & \dec{-29.8\%}\\
multi-agent & 20.58 & \dec{-21.7\%} & 19.17 & \dec{-38.9\%}\\
\midrule
descriptive & 34.55 & \dec{-17.7\%} & 33.58 & \dec{-22.8\%}\\
predictive & 24.75 & \dec{-18.5\%} & 24.3 & \dec{-33.5\%}\\
counterfactual & 41.3 & \dec{-1.6\%} & 40.4 & \dec{-15.8\%}\\
explanatory & 31.78 & \dec{-15.1\%} & 30.57 & \dec{-24.7\%}\\
\midrule
action & 15.72 & \inc{+4.6\%}  & 15.64 & \dec{-1.7\%}\\
object & 7.43 & \dec{-68\%} & 6.33 & \dec{-86.0\%}\\
state & 45.03 & \dec{-23.9\%} & 42.51 & \dec{-37.7\%}\\
change & 69.87 & \inc{+2.3\%} & 68.77 & \inc{+2.1\%}\\
\midrule
all & 33.92 & \dec{-10.6\%} & 32.51 & \dec{-23.0\%}\\
\bottomrule
\end{tabular}
}
\end{minipage}
\end{table}

\begin{table}[t!]
\caption{Model performance on the EgoTaskQA \textit{indirect} split.}
\label{tab:indirect_split}
\centering
\resizebox{\linewidth}{!}{%
\begin{tabular}{cccccccccc}
\toprule
 & Category & BERT & HCRN (w/o vision) & VisualBERT & PSAC & HME & HGA & HCRN & ClipBERT \\
\midrule
\multirow{3}{*}{\rotatebox[origin=c]{90}{Scope}} & world & 34.96 & 33.61 & 40.00 & \textbf{44.74} & 35.91 & 31.29 & 44.04 & 26.51\\
& intent & 23.56 & 23.98 & 36.02 & \textbf{48.38} & 31.73 & 20.42 & 47.02 & 14.66\\
& multi-agent & 19.70 & 19.25 & 26.02 & \textbf{35.37} & 25.07 & 17.74 & 30.11 & 20.09\\
\midrule
\multirow{4}{*}{\rotatebox[origin=c]{90}{Type}} & descriptive & 33.09 & 30.73 & 38.9 & \textbf{43.36} & 34.48 & 29.01 & 42.02 & 24.35\\
& predictive & 15.58 & 13.68 & 31.37 & 29.11 & 27.79 & 15.16 & \textbf{46.32} & 10.32\\
& counterfactual & 34.59 & 34.75 & 37.63 & 39.94 & 35.07 & 33.01 & \textbf{43.64} & 26.29\\
& explanatory & 27.38 & 28.11 & 32.75 & \textbf{42.53} & 29.16 & 24.00 & 39.69 & 22.46\\
\midrule
\multirow{4}{*}{\rotatebox[origin=c]{90}{Semantic}} & action & 26.91 & 28.18 & 27.49 & \textbf{30.06} & 25.12 & 26.15 & 29.61 & 25.25\\
& object & 2.808 & 4.13 & 22.63 & 30.97 & 19.08 & 7.02 & \textbf{32.20} & 10.49\\
& state & 21.96 & 21.24 & 32.02 & \textbf{43.29} & 31.60 & 17.67 & 41.81 & 15.29\\
& change & 55.28 & 50.71 & 55.59 & \textbf{57.20} & 47.65 & 47.22 & 56.27 & 35.26\\
\midrule
\multirow{3}{*}{\rotatebox[origin=c]{90}{Overall}} & open & 11.22 & 11.38 & 21.05 & \textbf{28.23} & 18.27 & 8.66 & 27.82 & 11.17\\
& binary & 58.24 & 55.52 & 57.61 & \textbf{60.30} & 52.55 & 53.72 & 59.29 & 40.71\\
& all & 31.78 & 30.76 & 37.01 & \textbf{42.25} & 33.06 & 28.36 & 41.56 & 24.08 \\
\midrule
\multicolumn{2}{c}{Performance Change} & \dec{-6.4\%} & \dec{-5.4\%} & \dec{-2.4\%} & \inc{+4.9\%} & \dec{-17.7\%} & \dec{-22.9\%} & \dec{-1.5\%} & \dec{-39.6\%}\\
% \midrule
% \multicolumn{2}{c}{Compared with \texttt{random}}& -6.4\% & -5.3\% & -2.4\% & +8.6\% & -17.7\% & -29.4\% & -0.1\% & -23.1\%\\
\bottomrule
\end{tabular}
}
\end{table}

\subsection{Generalizing to indirect references}\label{sec:exp:indirect}
On the EgoTaskQA \textit{indirect} split, we evaluate models' capability to leverage learned task knowledge for solving more complicated indirect reference tasks. With the \textit{normal} split allowing for shortcuts on action-state associations, the \textit{indirect} split forbids such exploitation by differentiating references during training and testing. As shown in~\cref{tab:indirect_split}, we observe more significant performance drops in language-only models compared to their vision-language counterparts. More specifically, the performance of BERT and language-only HCRN dropped 20.8\% and 26.3\% on the ``change'' category, where we observed potential exploitation on question texts in~\cref{sec:exp:text_modality}. This serves as a shred of evidence that the \textit{indirect} split helps reduce the possibility of exploiting simple associations in texts. As for baseline models, we recognize a common performance decrease shared by most models on the \textit{indirect} split. Among them, we notice a significant performance drop for ClipBERT, which conflicts with the dominating role of large-scale pretrained vision-language models on various reasoning tasks.
We suspect that this degeneration might originate from two lines of problems: (1) the model design on sampling fewer videos and aligning visual/text graphs directly, which conflicts with the intuition that detailed spatial-temporal information and reasoning is indispensable for grounding indirect references; and (2) adopting large-scale pre-trained models directly to a specific domain is non-trivial, especially with challenges in grounding knowledge to visual signals. Overall, our experiments on the EgoTaskQA \textit{indirect} split further reveals the demand for better spatial-temporal reasoning modules that solve the problem of compositional goal-oriented reasoning with indirect references.

% To further avoid mining text-based knowledge within questions, we design another indirect reference split on top of the EgoTaskQA questions. As discussed in~\cref{sec:data:answer_generation}, we generate indirect reference splits by aggregating basic indirect reference questions and other questions without indirect references as training data. In testing split, we leave only questions that contain both indirect references and the original reasoning logics. Such a design share the same spirit for exposing the uniqueness of spatial-temporal information for video question-answering. To answer these questions, models must first be able to ground over both actions and object state changes, and then based on these indirect references to provide answers for different reasoning desiderata. We provide the result of the six aforementioned video question reasoning models for the indirect split in~\cref{tab:indirect_split}.

% For ClipBERT, we suspect that the performance decrease comes from a negative effect of sparsely sampled video inputs, which might be detrimental for indirect reference understanding that requires accurate spatial-temporal localization. For HGA, we ascribe the performance decrease to the change of entities during training and testing that confused the graph alignment process. Both results conclude the demand for a good spatial-temporal reasoning module. 

\section{Conclusions}
\label{sec:conclusion}
We introduce the EgoTaskQA benchmark to systematically evaluate models' understanding of goal-oriented activities from an egocentric perspective. We annotate object states, relationships, and agents' beliefs on the LEMMA dataset. We generate diverse questions covering different reasoning capabilities and target the crucial dimensions of task understanding: action dependencies and effects, agents' intents and goals, and belief modeling. We evaluate state-of-the-art video question-answering models and show their gaps compared with the human on two challenging splits, \textit{normal} and \textit{indirect}, to promote future study on indirect reference understanding and goal-oriented reasoning.

\paragraph{Ethics} The EgoTaskQA benchmark is built upon LEMMA and contains different subjects. As noted in LEMMA, the authors obtain consent from subjects by signing an agreement form with potential impacts adequately informed before recording. We mainly annotated objective world status and multi-agent information using multiple-choice selection and asked annotators to annotate the awareness of other subjects' actions with binary options for subjective annotations. For the annotation process, the workers' agreements are obtained by the publicly available annotation service platform~\ac{amt} we adopted. In summary, we forbid subjective comments with no personally identifiable information revealed, and all participants' agreements are well addressed.

\paragraph{Limitations} Our work is primarily limited to two aspects. (i) Constrained by the data and annotation complexity, the scope of our activities is limited to indoor goal-oriented tasks. We believe that adding more diverse activities to EgoTaskQA will further increase its value as a general benchmark.
(ii) Although EgoTaskQA supports many more additional supervisions, we currently limit our discussions to the six state-of-the-art models to have a fair comparison. With the current result showing insights into problems and challenges in EgoTaskQA, we leave the exploration of model design to future work and briefly discuss the potential solutions as follows. Meanwhile, we will continue this data curation for a broader range of human activities. 

\paragraph{Future work} We plan to investigate the following two branches in the future, (i) explicit spatial-temporal grounding for modularized video QA models and (ii) prompting large-scale pre-trained models (both visual and language) for the domain-specific video QA challenges. Firstly, egocentric data can provide finer information and ease the challenge of grounding in modularized neuro-symbolic models. This could complement existing video reasoning methods and test the potential of neuro-symbolic models on complex reasoning tasks from a real-world, multi-agent, and causal perspective. Next, with increasing efforts in adapting large-scale pre-trained models for reasoning, our experiments suggest that adopting such models directly to a specific domain is non-trivial. Compared to their capabilities in commonsense reasoning, how to enable pre-trained models with the ability to fastly adapt to complex reasoning tasks still remains an interesting problem to be solved.

\paragraph{Broader impact} With most existing intelligent robots depending on the understanding of world states to act and plan, we hope that the augmented LEMMA dataset can bridge the study of world-model learning in simulated environments and real-world complex event understanding. Additionally, we believe the EgoTaskQA benchmark proposes challenges on goal-oriented reasoning and hope such efforts can foster research in broader video understanding directions, including video-language understanding, spatial-temporal grounding, task learning, future anticipation. We also see its potential in imitation learning and knowledge acquisition, which will further drive the study of intelligent assistive robots that can perform tasks coordinately with humans.

\paragraph{Public access} We host our videos, annotations, metadata, and question-answering pairs on our website. We provide videos in .mp4 format, metadata, annotations in JSON and pandas DataFrames, and question-answer pairs with their corresponding metadata in JSON format. We make our data publicly available under the CC BY-NC-SA license, which allows reusers to distribute, remix, adapt and build upon the material for noncommercial purposes only and only so long as attribution is given to the creator. We bear all responsibility in case of violation of rights.

\paragraph{Acknowledgement} We thank all colleagues from VCLA and BIGAI for fruitful discussions. We would also like to thank the anonymous reviewers for their constructive feedback. This work reported herein was supported by National Key R\&D Program of China (2021ZD0150200).

\bibliographystyle{unsrtnat}
\bibliography{ref}
\clearpage
\appendix

\section{Data Collection}\label{app:data}
\subsection{Data Statistics}\label{app:data:stats}
In this section, we provide more details and statistics on the annotated data. We leverage the compositional action annotation provided in LEMMA~\cite{jia2020lemma} and use objects in all semantic positions as the initial set of interacting objects. We remove temporally static, \ie not changing, objects like sofa, floor, hand, table and use them as general reference in action annotation (\eg we consider only ``bottled-water'' for action ``get \underline{bottled-water} from \underline{table} using {hand}'' and use ``get something from \underline{table}'' for action reference). We annotate states as well as spatial relationships of these objects for both before and after actions and obtain a total of 30K before-after pairs over the 10K action segments annotated in LEMMA. Moreover, we annotate the spatial relationships of the actor, his acting relationship (\eg ``get \underline{meat} using \underline{fork}'' indicates $\langle$P1$\rangle$[getting]$\langle$meat$\rangle$ and $\langle$P1$\rangle$[getting-with]$\langle$fork$\rangle$), as well as his multi-agent relationships annotated as discussed in~\cref{sec:data:annotation}.

\begin{figure}[!t]
    \begin{minipage}{\linewidth}
        \includegraphics[width=\linewidth]{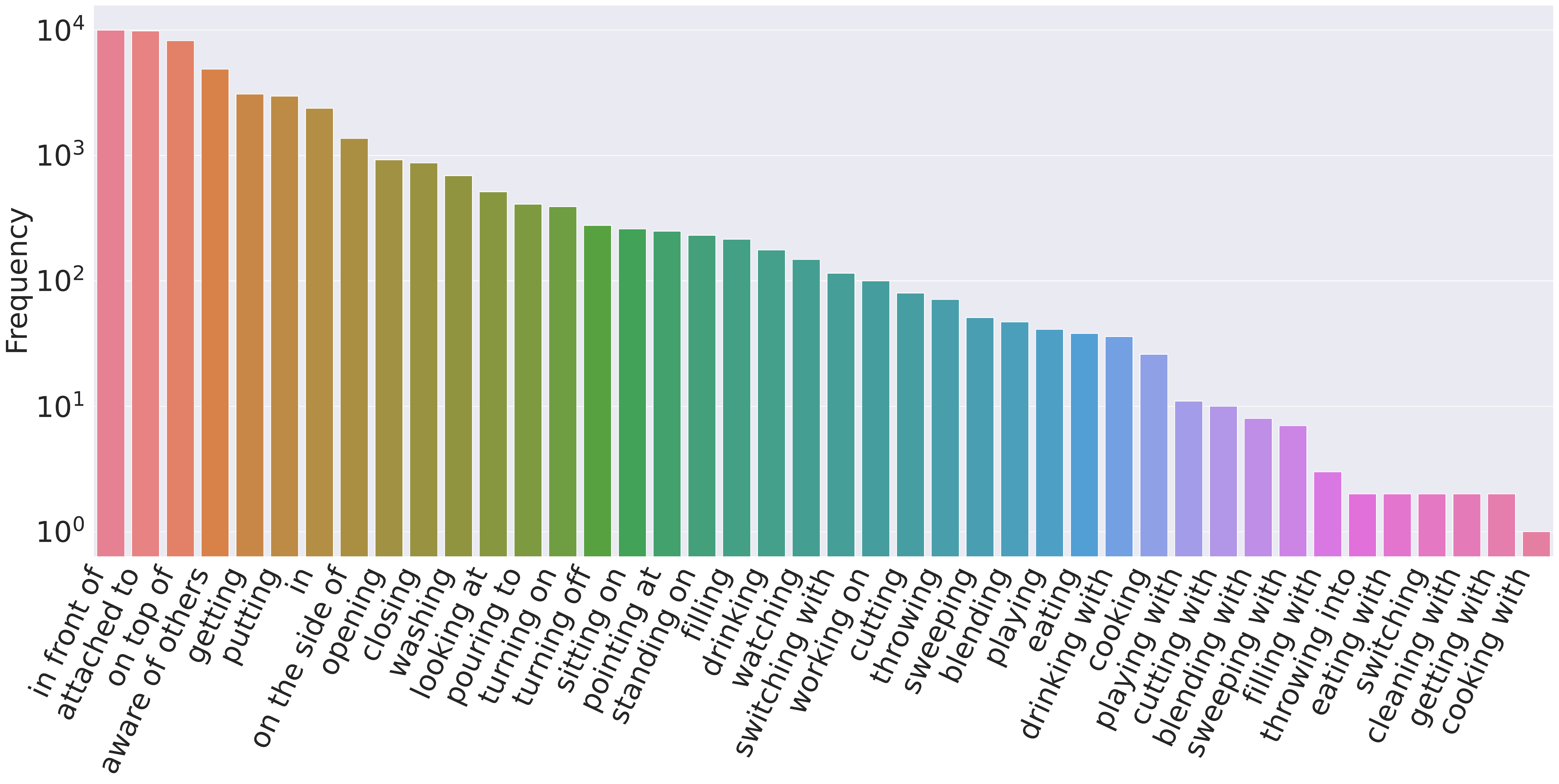}
        \captionof{figure}{Statistics of relationships annotated during EgoTaskQA data collection. Frequencies are normalized to log-scale for better visualization.}
        \label{fig:relationship_stats}
    \end{minipage} \\
    \begin{minipage}{\linewidth}
        \includegraphics[width=0.48\linewidth]{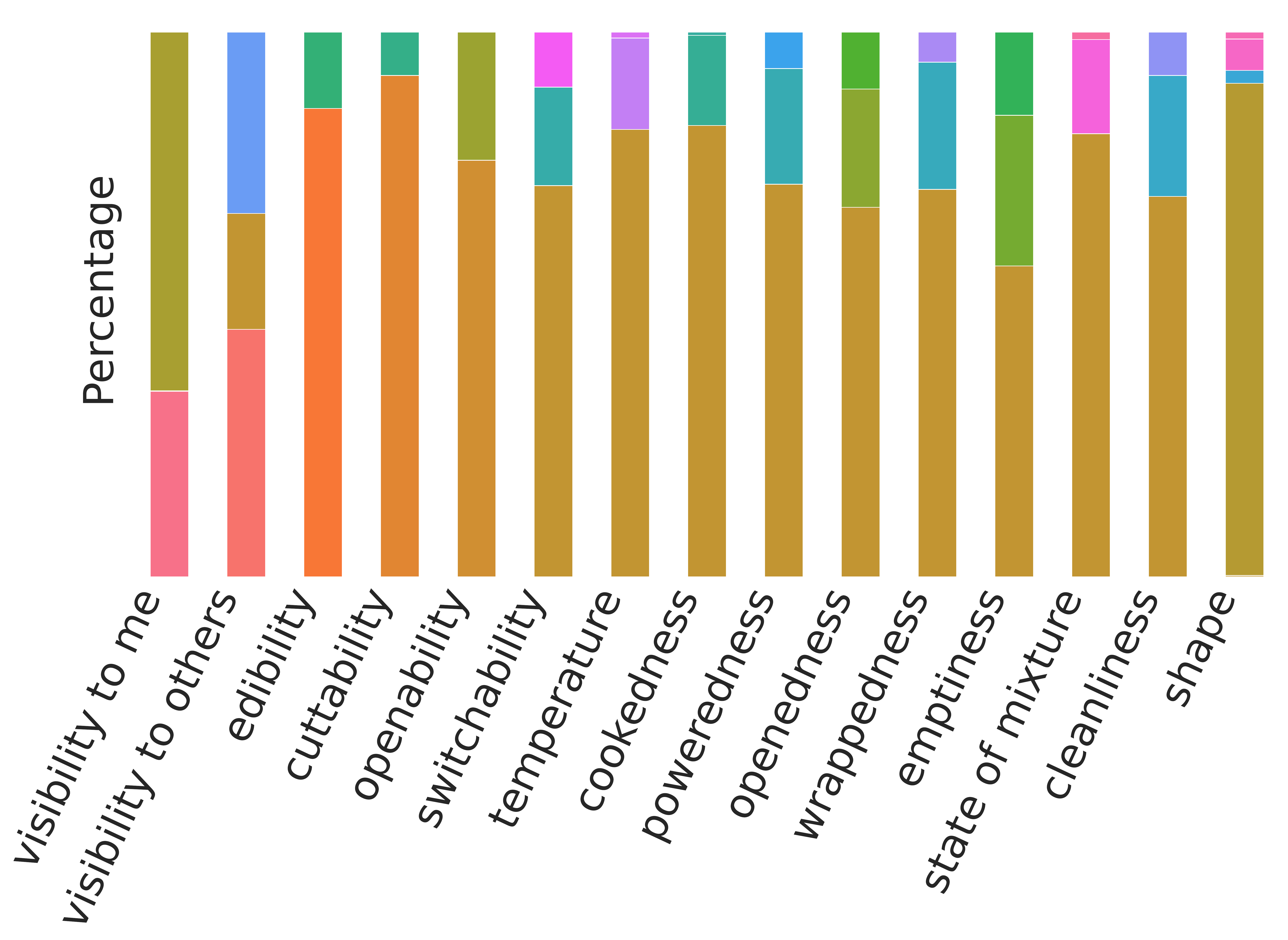}
        \hfill
        \includegraphics[width=0.52\linewidth]{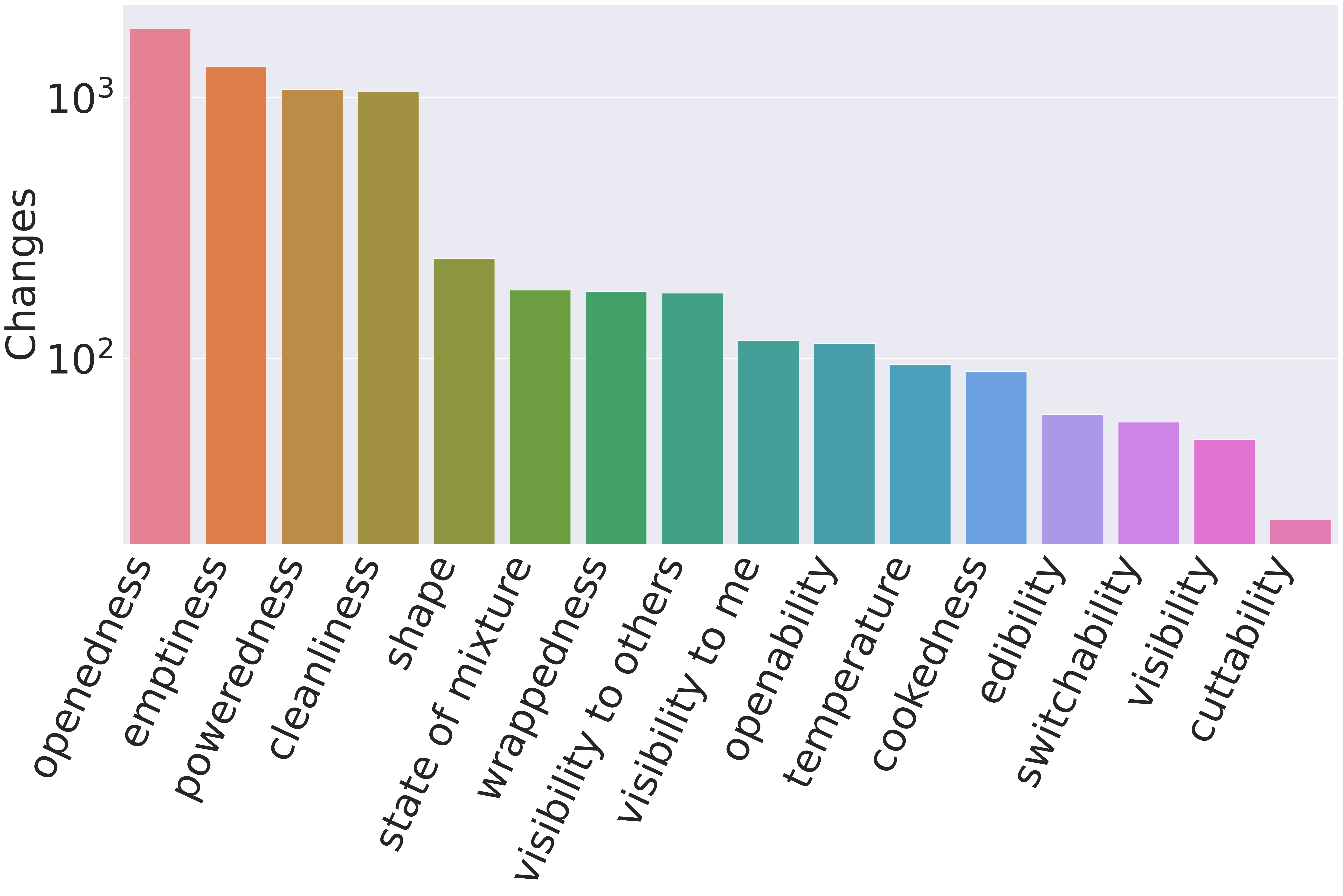}
        \captionof{figure}{Statistics of object state/attribute change: (left) the ratio of each annotated attribute value for its corresponding object attribute (\eg visible and invisible for ``visibility to me''); (right) the frequency of object attribute that was changed by actions (normalized to log-scale for better visualization).}
        \label{fig:state_change_stats}
    \end{minipage}
\end{figure}

\begin{figure}
    \centering
    \includegraphics[width=0.95\linewidth]{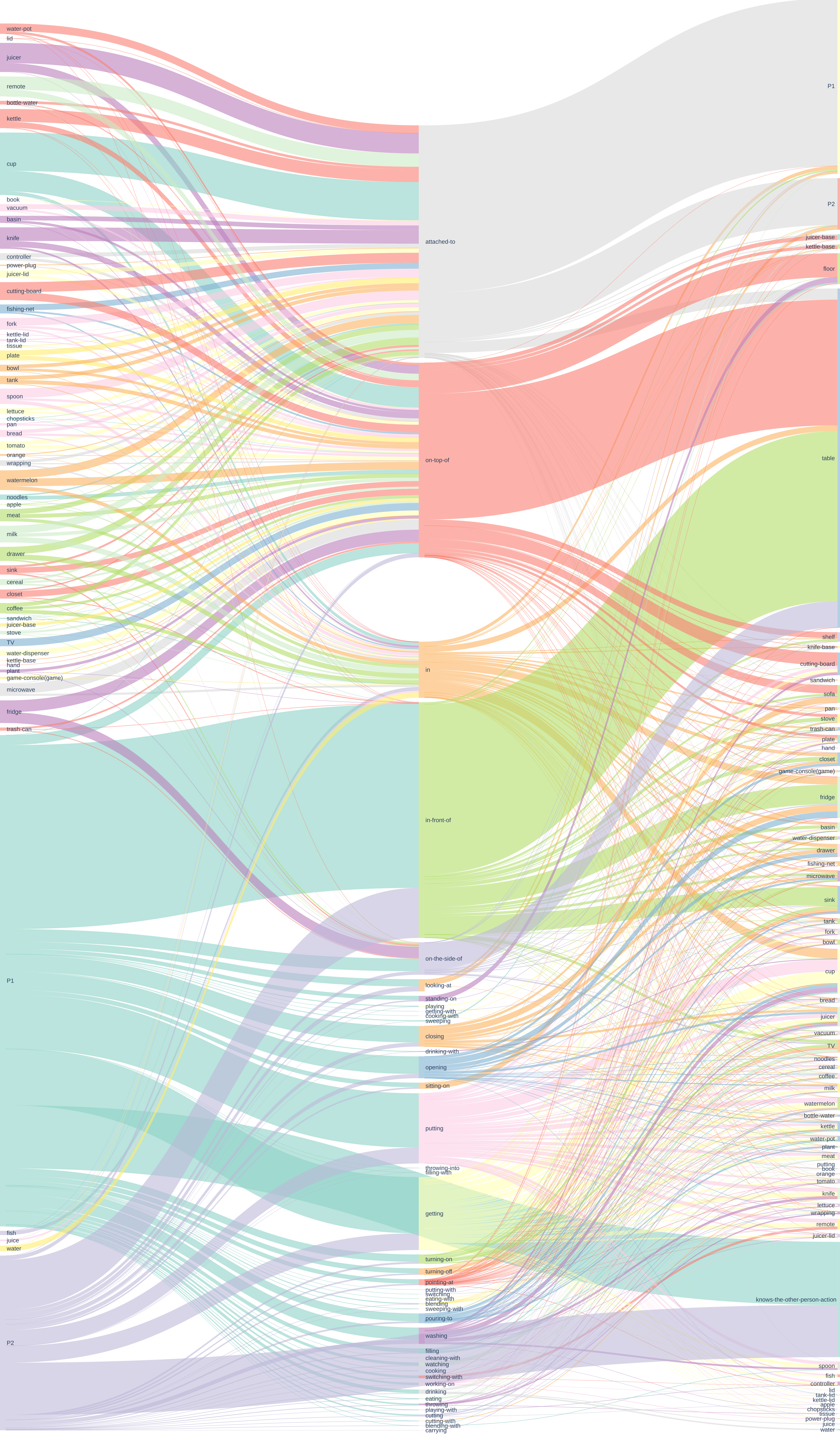}
    \caption{Visualization of all relationship pairs and their corresponding ratios.}
    \label{fig:relationship_sankey}
\end{figure}

We visualize the statistics of annotated relationships in~\cref{fig:relationship_stats}. As we can see from the histogram, spatial relationships of objects were annotated the most, followed by multi-agent relationships like ``aware of others'' and ``looking at''. This meets our expectation of the frequent changes in objects' spatial relationships during goal-oriented task execution. Action-related relationships also make up a considerable portion of overall relationship annotations and describe detailed relationships between the person and the target object (\eg getting, putting, pouring) or the tool object (\eg getting-with, cutting-with, putting-with). We visualize the statistics of relationship pairs in~\cref{fig:relationship_sankey}.

We list all annotated object attributes and their state values in~\cref{tab:fluent_val}, and visualize their statistics in~\cref{fig:state_change_stats}. We add an option ``unknown'' to all attributes for annotating unclear scenarios and ignore this answer during question generation. As shown in~\cref{tab:fluent_val} and~\cref{fig:state_change_stats}, we consider various time-varying object attributes including visibility, affordance (\eg cuttability, edibility), and task-dependent status (\eg emptiness, shape). In~\cref{fig:state_change_stats} (right), we plot the number of changes for each object attribute. In addition to spatial relationship changes described previously, there is an increasing number of occurrences from affordance changes to visibility changes and, finally, task-dependent status changes. As LEMMA is recorded in indoor environments (kitchens and living rooms), we also observe a large amount of containment relationship changes (``open/close'' and ``emptiness''). We argue that this data is also potentially beneficial for the study of containment relationships~\cite{liang2016inferring}.

\begin{table}[!h]
    \caption{A full list of time-varying object attributes considered and their corresponding possible values.}
    \label{tab:fluent_val}
    \centering
    \resizebox{\linewidth}{!}{
        \begin{tabular}{ccc}
        \toprule
        Attribute & Type & Possible State Values \\
        \midrule
        visibility to me & visibility & visible to me / invisible to me / unknown\\
        visibility to the other person& visibility  & visible to the other person / invisible to the other person / unknown\\
        edibility & affordance  & edible / can not be eaten / unknown\\
        cuttability & affordance & cuttable / not cuttable / unknown \\
        openability & affordance & openable / can not be opened / unknown \\
        switchability & affordance & can be turned on / can not be turned on / unknown \\
        temperature & status & boiled / in room temperature / unknown \\
        poweredness & status & on / off / unknown \\
        cookedness & status & cooked / raw / unknown \\
        wrappedness & status & wrapped / unwrapped / unknown \\
        emptiness & status & empty / full / unknown \\
        state of mixture & status & mixing / not mixing / unknown \\
        cleanliness & status & clean / dirty / unknown \\
        shape & status & whole / part / diced / fluid / unknown\\
        \bottomrule
        \end{tabular}
    }
\end{table}

\subsection{Causal Dependency}\label{app:data:causal}

In this section, we describe the details for determining the causal relationship between actions. We adopt rules for deciding the causal dependency between actions. Given two actions $a_1$ and $a_2$, and their state annotations $s_1$ and $s_2$, we determine the causal dependency between them as shown in~\cref{alg:causal_dependency}. 

We first collect all interactive object set $O_1$ and $O_2$ for $a_1$ and $a_2$, and see if there exists an overlap of objects. 
If no, we assume $a_1$ and $a_2$ is not \textit{related}.
Next, for each object $o$ that is interacted in both actions, 
we check whether $a_1$ lead to the change of attribute $s$, 
\begin{wrapfigure}[20]{r}{0.48\linewidth}
\begin{minipage}{\linewidth}
\resizebox{\linewidth}{!}{
    \begin{algorithm}[H]
    \caption{Causal Dependency Check}
    \label{alg:causal_dependency}
    \DontPrintSemicolon
    \KwIn{two actions $a_1$ and $a_2$ and their object state annotation $S_1$ and $S_2$.}
    \KwOut{the causal dependency relationships between $a_1$ and $a_2$.}
    Gather all interactive objects $O_1=\{o_i^1\}_{i=1}^{m}$ and $O_2=\{o_i^2\}_{i=1}^{n}$ in action $a_1$ and $a_2$. \\
    \uIf{$O_1 \cap O_2 = \varnothing$}{
        \Return{unrelated}
    }
    \lElse{
        \For{$o \in O_1 \cap O_2$}{
            \For{$s_{1,o}\in S_1$, $s_{2,o}\in S_2$}{
                \uIf{$(s_{1,o}^{\text{before}} \neq s_{1, o}^{\text{after}})\land$$(s_{1,o}^{\text{after}} = s_{2,o}^{\text{before}})$ $\land(s_{2,o}^{\text{before}} \neq s_{2,o}^{\text{after}})$}{
                    \Return{dependent}
                }
                \lElseIf{$(s_{1,o}^{\text{before}} \neq s_{1, o}^{\text{after}})\land$
                $(s_{1,o}^{\text{after}} = s_{2,o}^{\text{before}})$}{\Return{related}}
            }
        }
        \Return{unrelated}
    }
    \end{algorithm}
}
\end{minipage}
\end{wrapfigure}
which is a precondition of $a_2$'s change on $o$. 
This condition is validated by checking if $o$ changed the same attribute $s$ in both $a_1$ and $a_2$, and the status after $a_1$ equals the status before $a_2$, \ie $s_{1,o}^{\text{after}} = s_{2,o}^{\text{before}}$. 
We say that $a_1$ and $a_2$ are causally \textit{dependent} if this condition is satisfied. If there exists an attribute $s$ that was affected by $a_1$ and did not change during $a_2$, \ie $(s_{1,o}^{\text{before}} \neq s_{1, o}^{\text{after}})\land(s_{1,o}^{\text{after}} = s_{2,o}^{\text{before}})$, we say that these two actions are \textit{related} since we can not determine whether this relationship is causal or not from the annotations. As currently we did not use additional human  resources for verifying each of this \textit{related} actions, we limit our scope of question generation to the \textit{dependent} and \textit{unrelated} action pairs. After checking the causal dependency for all action pairs in the video, we recursively construct the dependency tree by taking each action as root and adding actions that are dependent on all dependants of the action to its dependants set. During the recursion, we update the dependency for a newly added action to \textit{related} if there exist \textit{related} dependency relationships in the path from the root action to it.

\section{Question-Answer Generation}

\subsection{Preprocessing Annotations}
To generate answers, we first collect video intervals as mentioned in~\cref{sec:data:annotation}. These clips are cropped from original videos to contain 4$\sim$5 actions on average. We further concatenate the next three actions performed by the actor that is unseen in videos into the intervals of interest for generating predictive questions. After generating these intervals of interest, we gather all corresponding annotations, including annotations for both the actor's action and the helper's (\ie the other person's) simultaneous actions. We organize these annotations in a dictionary for convenience purposes.
\subsection{Operator and Program Design}
We use a template-based method for generating questions. More specifically, we design operators that work on the annotation dictionaries with different purposes. Inspired by previous works~\cite{johnson2017clevr,yi2020clevrer,grunde-mcLaughlin2021agqa}, we design nine basic operators for composing the logic for each program template. We provide the specification of each operator, and its usage with an example, in~\cref{tab:operations}. The basis of these programs lies in the conditional query, similar to database queries. We use $A@B$ for filtering data with the attribute $A$ equal $B$, and we use $A\$$ for querying the value of attribute $A$ from data. 
\begin{table}[h!]
    \caption{All program modules used for question-answer generation.}
    \label{tab:operations}
    \resizebox{\linewidth}{!}{
        \renewcommand{\arraystretch}{1}
        \begin{tabular}{clccc}
        \toprule
        Query Operation & Parameter List & Return Type & Usage and Example\\
        \midrule
        
        \multirow{2}{*}{Filter} & arg$_1$: conditions & \multirow{2}{*}{intervals} & Return the intervals that satisfies the conditions.\\
        & arg$_2$: intervals & & \texttt{filter([obj@spoon, change@cleanliness], video)
        }\\
        
        \multirow{2}{*}{Only} & \multirow{2}{*}{arg$_1$: intervals} & \multirow{2}{*}{interval} & Return the only interval from list, return \texttt{None} if $|\text{arg}_1| \neq 1$.\\
        & & & \texttt{only(filter([action@putting]))}\\
        
        \multirow{2}{*}{Localize} & arg$_1$: before/after & \multirow{2}{*}{intervals} & Return all intervals before/after the interval provided in arg$_2$.\\
        & arg$_2$: interval & & \texttt{localize(before, only(filter(action@putting)))} \\
        
        \multirow{2}{*}{IterateUntil} & arg$_1$:forward/backward & \multirow{2}{*}{interval} & Return the first interval of the interval list from the front/back.\\
        & arg$_2$: intervals & & \texttt{iterate\_until(forward, filter([change@emptiness], video))}\\
        
        \multirow{2}{*}{Query} & arg$_1$: conditions & \multirow{2}{*}{value} & Return the value from the interval identified by the conditions. \\
        & arg$_2$: interval & & \texttt{query([aware@yes, action\$], only(filter([action@getting], video)))} \\
        \multirow{2}{*}{Verify} & arg$_1$: conditions & \multirow{2}{*}{bool} & Verify arg$_1$ in the interval arg$_2$, return ``yes'' if satisfied, ``no'' otherwise.\\
        & arg$_2$: interval & & \texttt{verify([change@openedness, obj@closet], only(filter([action@closing], video)))}\\
        \multirow{2}{*}{Pred} & \multirow{2}{*}{arg$_1$: intervals} & intervals & Return the anticipating intervals within intervals arg$_1$\\
        & & & \texttt{pred(filter([action@pouring], video))}\\
        \multirow{2}{*}{Counterfactual} & arg$_1$: conditions & intervals & Return the original intervals with executability of each interval adjusted according to the counterfactual query arg$_2$.\\
        & arg$_2$:intervals & & \texttt{counterfactual([action@getting], video)}\\
        \multirow{2}{*}{Depend} & arg$_1$: interval & \multirow{2}{*}{bool} & Return ``yes'' if interval arg$_1$ and interval$_2$ are dependent, ``no'' otherwise. \\
        & arg$_2$: interval & & \texttt{depend(only(filter([action@opening], video)), only(filter([action@closing], video)))} \\
        \bottomrule
        \end{tabular}
    }
\end{table}

We provide the full list of program templates in~\cref{tab:question_templates}. In these templates, we use \{$a$\} for representing the parameter type action, \{$o$\} for the parameter type object, \{$f$\} for object attributes and \{$fv$\} for attribute values. During the question generation process, we substitute these positional arguments with the corresponding sample space to initialize these program templates so that the resulting programs are executable on the annotation dictionary. If the program becomes not executable at any intermediate step, we return \texttt{None} for the corresponding operator to stop the trial, reinitialize the template and make a new attempt. This generation process concludes to 368K exhaustively searched questions that have answers.

For indirect questions, we make use of the provided templates and use indirect reference to substitute parameters for action (\{$a$\}) and object (\{$o$\}). We list all indirect reference considered in~\cref{tab:indirect_reference}. In our experiments, we consider simple temporal references of actions, including ``before some action'', ``after some action'', ``the first action'' and ``the last action''. For object references, we consider using state change for referring objects. We use ``the first object that has status change'', ``the last object that has status change'' as well as ``the object that has status change'' to refer to objects. Concretely, we substitute the corresponding positional arguments \{$a$\} and \{$o$\} shown in~\cref{tab:question_templates} with the substituting text and program templates shown in in~\cref{tab:indirect_reference}. We use the template with $\langle$ and $\rangle$ to substitute the action/object positional arguments in the original program template for generating programs. As this substitution could be easily adapted to have multi-step indirect references, we limit the indirect references in our benchmark to 1-step indirect references to avoid generating questions that are difficult to understand. To facilitate models' understanding of indirect references, we add additional questions on these indirect queries for objects and actions. We leverage the program template shown in~\cref{tab:indirect_reference} and adjust the text to have questions like ``what is the first action...'', ``what is the action before/after...'', and ``which object changed its status first..''.

\begin{table}[b!]
\caption{Question-answer pair statistics before and after balancing.}
\label{tab:balance_statistics}
\resizebox{\linewidth}{!}{
    \begin{tabular}{cccccccccccccc}
    \toprule
     & World & Intent & Multi-agent & Descriptive & Predictive & Counterfactual & Explanatory & Action & Object & State & Change & Open & Binary \\
     \midrule
    Before & 299K  & 43K & 53K & 181K & 22K & 71K & 102K & 122K & 14K & 105K & 126K & 182K & 186K \\
    After & 32K & 5K & 6K & 21K & 4K & 6K & 9K & 17K & 4K & 9K & 10K & 26K & 13K \\
    \bottomrule
    \end{tabular}
}
\end{table}

\begin{sidewaystable}[ph!]
    \centering
    \begin{minipage}{\linewidth}
        \caption{Question templates adopted in EgoTaskQA. We use ``obj'' short for ``object'' and ``hoi'' short for ``human-object interaction''.}
        \label{tab:question_templates}
        \resizebox{\linewidth}{!}{
        \begin{tabular}{cccccc}
        \toprule
Template & Program & Type & Scope & Semantic & Overall \\
\midrule
which object changed its status when the person \{$a$\}? & \texttt{query({[}change, obj\${]}, only(filter({[}action@\{$a$\}{]}, video)))} & descriptive & world & object & query \\
what status of \{$o$\} changed while the person \{$a$\}? & \texttt{query({[}change, obj@\{$o$\}, change\${]}, only(filter({[}action@\{$a$\}{]}, video)))} & descriptive & world & change & query \\
Is \{$o$\} visible to the other person \{t\} the person \{$a$\}? & \texttt{query({[}state, type@\{t\}, obj@\{$o$\}, 'visibility to the other person'\${]}, only(filter({[}action@\{$a$\},is\_multi@yes{]}, video)))} & descriptive & world, multi-agent & state & verify \\
Is the other person aware when the person \{$a$\}? & \texttt{query({[}aware{]}, only(filter({[}others{]}, filter({[}action@\{$a$\}{]}, video))))} & descriptive & multi-agent & action & verify \\
what is the other person doing while the person \{$a$\}? & \texttt{query({[}hoi{]}, only(filter({[}aware@yes, others{]}, filter({[}action@\{$a$\}{]}, video))))} & descriptive & multi-agent & action & query \\
which object changed its status when the other person \{$a$\}? & \texttt{query({[}change, obj\${]}, only(filter({[}action@\{$a$\}{]}, filter({[}aware@yes, others{]}, video))))} & descriptive & world, multi-agent & object & query \\
What is the status of \{$o$\} \{t\} the person \{$a$\} to change it? & \texttt{query({[}change, obj@\{$o$\}, \{t\}\${]}, only(filter({[}action@\{$a$\}{]}, video)))} & descriptive & world & state & query \\
what is the status of \{$o$\} \{t\} the other person \{$a$\} to change it? & \texttt{query({[}change, obj@\{$o$\}, \{t\}\${]}, only(filter({[}action@\{$a$\}{]}, filter({[}aware@yes, others{]}, video))))} & descriptive & world, multi-agent & state & query \\
did the attribute of \{$o$\} changed because of the person's action \{$a$\}? & \texttt{verify({[}change, obj@\{$o$\}{]}, only(filter({[}action@\{$a$\}{]}, video)))} & descriptive & world & change & verify \\
during which action does the person knows about the other person's action? & \texttt{query({[}hoi{]}, only(filter({[}aware@yes{]}, video)))} & descriptive & multi-agent & action & query \\
what will the person want to have \{$o$\}'s \{$f$\} be in the future? & \texttt{query({[}change, obj@\{$o$\}, change@\{$f$\}, after\${]}, only(filter({[}change, obj@\{$o$\}, change@\{$f$\}{]}, pred(video))))} & predictive & intent & state & query \\
what status will the person change on \{$o$\}? & \texttt{query({[}change, obj@\{$o$\}, change\${]}, only(filter({[}change, obj@\{$o$\}{]}, pred(video))))} & predictive & intent & change & query \\
what does the other person want to have the \{$f$\} of \{$o$\} be? & \texttt{query({[}change, obj@\{$o$\}, change@\{$f$\}, after\${]}, only(filter({[}change, obj@\{$o$\}, change@\{$f$\}{]}, pred(filter({[}others{]}, video)))))} & predictive & intent & state & query \\
what will the other person change on \{$o$\}? & \texttt{query({[}change, obj@\{$o$\}, change\${]}, only(filter({[}change, obj@\{$o$\}{]}, pred(filter({[}others{]}, all)))))} & predictive & intent & state & query \\
what will the person do next after this video? & \texttt{query({[}hoi{]}, iterate\_until(forward, pred(video)))} & predictive & intent & action & query \\
what will the other person do next? & \texttt{query({[}hoi{]}, iterate\_until(forward, filter({[}aware@yes, others{]}, pred(video))))} & predictive & intent, multi-agent & action & query \\
will \{$o$\} be visible to the other person after the person's next action? & \texttt{query({[}state, type@after, obj@\{$o$\}, 'visibility to the other person'\${]}, iterate\_until(forward, filter({[}is\_multi@yes{]}, pred(video))))} & predictive & world, intent, multiagent & state & verify \\
if the actor do not \{$a$\}, which object will he/she not be able to change in the future? & \texttt{query({[}change, obj\${]}, only(filter({[}executable@no{]}, filter({[}change{]}, pred(counterfactual({[}action@\{$a$\}{]}, all))))))} & predictive, counterfactual & world & object & verify \\
what will the status of \{$o$\} change to if the actor \{$a$\} in the future? & \texttt{query({[}change, obj@\{$o$\}, after\${]}, only(filter({[}action@\{$a$\}{]}, pred(video))))} & predictive, counterfactual & world & state & query \\
which object's status is possible to be changed in the future? & \texttt{query({[}change, obj\${]}, only(filter({[}change{]}, pred(video))))} & predictive & world & object & query \\
if the person did not \{$a_1$\}, is the person able to \{$a_2$\}? & \texttt{query({[}executable{]}, only(filter({[}action@\{$a_2$\}{]}, counterfactual({[}action@\{$a_1$\}{]}, all))))} & counterfactual & world & action & verify \\
if the person did not \{$a$\}, what remaining actions in the video is executable? & \texttt{query({[}hoi{]}, only(filter({[}executable@yes{]}, counterfactual({[}action@\{$a$\}{]}, all))))} & counterfactual & world & action & query \\
if the person did not \{$a$\}, what remaining actions in the video is not executable? & \texttt{query({[}hoi{]}, only(filter({[}executable@no{]}, counterfactual({[}action@\{$a$\}{]}, all))))} & counterfactual & world & action & query \\
if the other person did not \{$a_1$\}, is the person able to \{$a_2$\}? & \texttt{query({[}executable{]}, only(filter({[}action@\{$a_2$\}{]}, counterfactual({[}others, action@\{$a_1$\}{]}, all))))} & counterfactual & multi-agent & action & verify \\
if the other person did not \{$a$\}, what actions of this person in the video is executable? & \texttt{query({[}hoi{]}, only(filter({[}executable@yes{]}, counterfactual({[}others, action@\{$a$\}{]}, all))))} & counterfactual & multi-agent & action & query \\
if the other person did not \{$a$\}, what actions of this person in the video is not executable? & \texttt{query({[}hoi{]}, only(filter({[}executable@no{]}, counterfactual({[}others, action@\{$a$\}{]}, all))))} & counterfactual & multi-agent & action & query \\
if the person did not \{$a$\}, will \{$o$\} change its status? & \texttt{query({[}executable{]}, only(filter({[}change, obj@\{$o$\}{]}, counterfactual({[}action@\{$a$\}{]}, all))))} & counterfactual & world & change & verify \\
what does the person want \{$o$\} to be for doing the action \{$a$\} in the video? & \texttt{query({[}change, obj@\{$o$\},after\${]}, only(filter({[}action@\{$a$\}{]}, video)))} & explanatory & intent & state & query \\
which attribute does the person want to change with \{$o$\} for doing the action \{$a$\} in the video? & \texttt{query({[}change, obj@\{$o$\}, change\${]}, only(filter({[}action@\{$a$\}{]}, video)))} & explanatory & intent & change & query \\
how did the person changed the \{$f$\} of \{$o$\}? & \texttt{query({[}hoi{]}, only(filter({[}change, obj@\{$o$\}, change@\{$f$\}{]}, video)))} & explanatory & world & action & query \\
what action caused \{$o$\}'s status to change to \{$fv$\}? & \texttt{query({[}hoi{]}, only(filter({[}change, obj@\{$o$\}, change@\{$f$\}, after@\{$fv$\}{]}, video)))} & explanatory & world & action & query \\
does action \{$a_1$\} fulfills the preconditions of action \{$a_2$\}? & \texttt{depend(only(filter({[}action@\{$a_1$\}{]}, video)), only(filter({[}action@\{$a_2$\}{]}, video)))} & explanatory & world & action & verify \\
how would \{$a$\} change the state of \{$o$\}? & \texttt{query({[}change, obj@\{$o$\}, after\${]}, only(filter({[}action@\{$a$\}{]}, video)))} & explanatory & world & state & query \\
what is the precondition of changing the \{$f$\} of \{$o$\}? & \texttt{query({[}change, change@\{$f$\}, obj@\{$o$\}, before\${]}, only(filter({[}change, change@\{$f$\}, obj@\{$o$\}{]}, video)))} & explanatory & world & state & query \\
is the person's action of \{$a_1$\} depending on the other person's action \{$a_2$\}? & \texttt{depend(only(filter({[}action@\{$a_2$\}{]}, filter({[}others{]}, video))), only(filter({[}action@\{$a_1$\}{]}, video)))} & explanatory & multi-agent & action & verify \\
\bottomrule
\end{tabular}
}
    \end{minipage}
    \\
    \begin{minipage}{\linewidth}
    \caption{Indirect references to objects and templates.}
    \label{tab:indirect_reference}
    \resizebox{\linewidth}{!}{
    \begin{tabular}{ccc}
    \toprule
    \multirow{2}[2]{*}{Type} & \multicolumn{2}{c}{Indirect Reference} \\
     & Text & Program \\
     \midrule
    \multirow{4}{*}{action} & the action after he/she \{$a$\} & $\langle$\texttt{query({[}action{]}, iterate\_until(forward, localize(after, only(filter({[}action@\{a\}{]}, video)), filter({[}{]}, video))))}$\rangle$ \\
     & the action before he/she \{$a$\} & $\langle$\texttt{query({[}action{]}, iterate\_until(backward, localize(before, only(filter({[}action@\{a\}{]}, video)), filter({[}{]}, video))))}$\rangle$ \\
     & the first action in the intervals & $\langle$\texttt{query({[}action{]}, iterate\_until(forward, filter({[}{]}, video)))}$\rangle$ \\
     & the last action in the intervals & $\langle$\texttt{query({[}action{]}, iterate\_until(backward, filter({[}{]}, video)))}$\rangle$ \\
     \midrule
    \multirow{3}{*}{object} & the object that has status change & $\langle$\texttt{query({[}change, obj\$], only(filter([change, obj\${]}, video)))}$\rangle$ \\
     & the first object that has status change & $\langle$\texttt{query({[}change, obj\$], iterate\_until(forward, filter([change, obj\${]}, video)))}$\rangle$ \\
     & the last object that has status change & $\langle$\texttt{query({[}change, obj\$], iterate\_until(backward, filter([change, obj\${]}, video)))}$\rangle$ \\
     \bottomrule
    \end{tabular}
    }
    \end{minipage}
\end{sidewaystable} 

\subsection{Question-answer statistics and balancing}
\begin{figure}[t!]
    \centering
    \resizebox{0.85\linewidth}{!}{
        \includegraphics{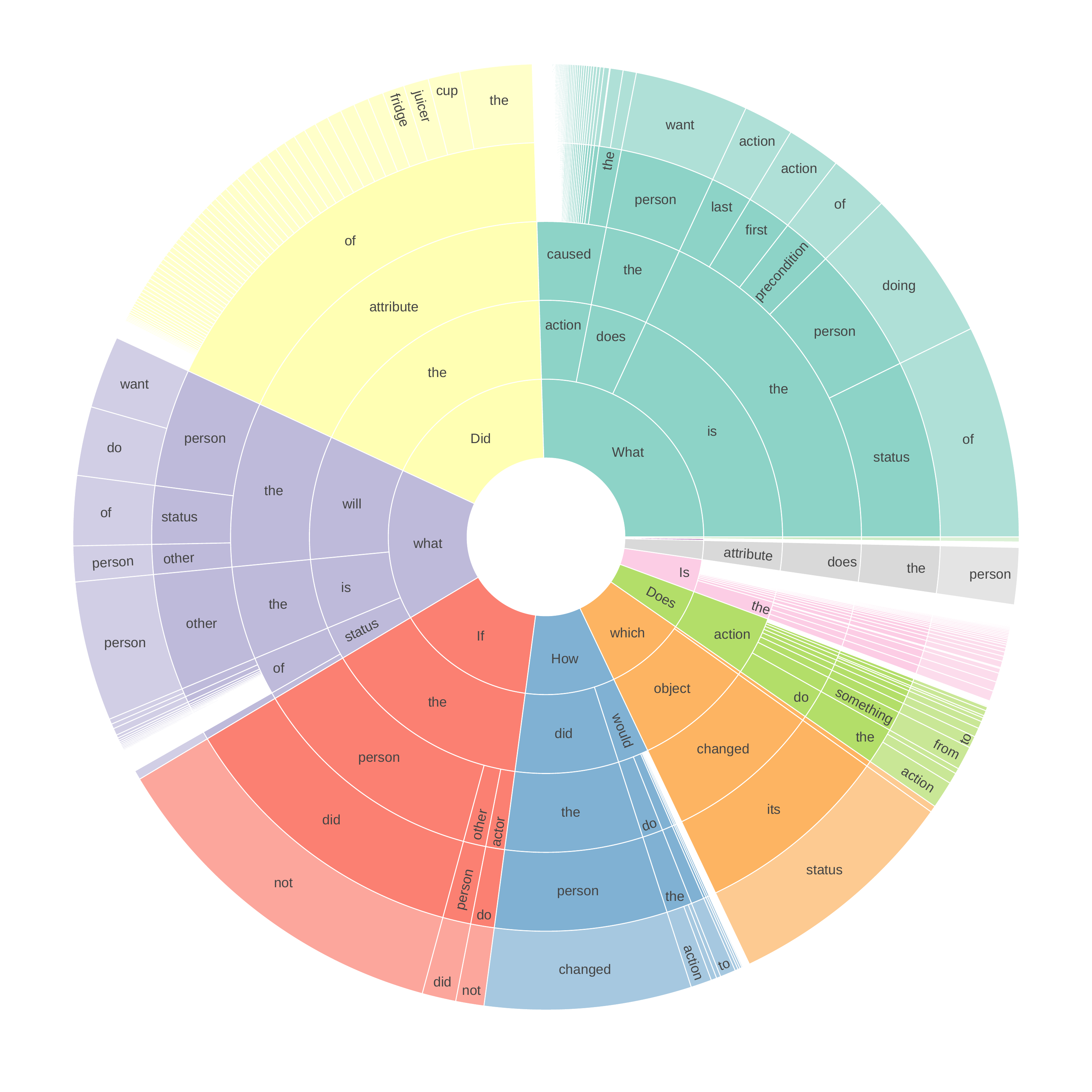}
    }
    \caption{Questions in EgoTaskQA targets different aspects of goal-oriented reasoning with various questions.}
    \label{fig:qa_sunburst}
\end{figure}

In this section, we provide the details for balancing and question-answer statistics. As described in~\cref{sec:qa:generation}, we balance the questions according to their reasoning types and obtain 40K diverse question-answer pairs. We visualize the most common question texts in~\cref{fig:qa_sunburst}. For balancing, we follow the algorithm provided by~\cite{grunde-mcLaughlin2021agqa} and adjust the open-answer problems to ensure that the top 20\% answers of each reasoning type do not answer to more than 33\% questions in the same type. We select this ratio to get a smoother answer distribution while not deleting too many questions in the whole set. To avoid overfitting to the binary answer distribution, we control the ratio between open-answer and binary questions to be 2:1. We show the statistics for each general question type before and after balancing in~\cref{tab:balance_statistics}.

\pagebreak

\section{Experiment}
In this section, we provide details on model implementation, hyper-parameters selection, and environment setup. We provide the details for each evaluated baseline model as follows:

\begin{itemize}[leftmargin=*]
\item \textbf{VisualBERT}: We use the pretrained VisualBERT model and implementation provided by Hugging Face~\cite{visualbert_huggingface}. Specifically, we uniformly sample 20 frames per video and extract visual features using ResNet to generate visual tokens. We use the pretrained BertTokenizer for embedding text tokens. To avoid instabilities during training, we set the learning rate to $5\times10^{-5}$. As we observe convergence at around 35 epochs, we set the total training epochs to be 40 with a batch size of 32. For its language-only variant, BERT, we adopt the same setting and train for 25 epochs as we observe faster convergence compared to its vision-language counterpart.

\item \textbf{HGA}: We follow the model setting provided by~\cite{jiang2020reasoning}. More specifically, We sample 20 frames per video uniformly and extract appearance features with VGG. For motion features, we sample 8 clips from each video and sample 16 frames from each clip to extract motion features with C3D~\cite{tran2015learning}. We embed the text with Glove~\cite{pennington2014glove} embeddings to generate text tokens. We adopt the experimental configuration designed for Frame-QA on T-GIF~\cite{jang2017tgif-qa} and set the learning rate to $1\times10^{-4}$. We train the model for 100 epochs with a batch size of 64.

\item \textbf{HME}: We follow the model setting provided by~\cite{fan2019heterogeneous}. As done similarly in HGA, we preprocess visual inputs with VGG and C3D for appearance and motion features and embed textual inputs with Glove embeddings. We adopt the experimental configuration designed for MSRVTT-QA~\cite{xu2017video}. We set the learning rate to $1\times10^{-3}$ and train the model for 25 epochs with a batch size of 32.

\item \textbf{PSAC}: We follow the model setting provided by~\cite{li2019beyond}. We uniformly sample 20 frames per video and extract appearance features using ResNet for visual input and embed textual inputs with Glove embeddings. We adopt the experimental configuration designed for Frame-QA on T-GIF. We set the learning rate to $1\times10^{-3}$ and train the model for 50 epochs with a batch size of 32.

\item \textbf{HCRN}: We follow the model setting provided by~\cite{le2020hierarchical}. We divide every video into eight equal length clips. Each clip is consisted of 16 frames and is used for obtaining two sources of information: frame-wise appearance feature extracted by ResNet, and motion feature extracted by ResNeXt. Textual inputs are embedded with Glove embeddings. We adopt the experimental configuration designed for MSRVTT-QA~\cite{xu2017video}. We set the learning rate to $1\times10^{-4}$ and train the model for 50 epochs with a batch size of 32.

\item \textbf{ClipBERT}: We follow the model setting provided by~\cite{lei2021less}. We preprocess the videos and texts into the ClipBERT format and adopt the experimental configuration designed for MSRVTT-QA~\cite{xu2017video}. As we have found instabilities during training (\ie NaNs and Infs in gradients), we reduce the learning rate for both transformers and CNN to $2\times10^{-5}$. We train the ClipBERT model for 25 epochs with a batch size of 24.
\end{itemize}

To provide a clear picture of all experimental settings and hyperparameters selected, we list training information for each model in~\cref{tab:hyper_parameter}.
We run all experiments on a single NVIDIA A100 (80G) GPU.
We provide all codes, checkpoints, and instructions for reproducing the experiments on our website.
\begin{table}[h!]
\caption{Hyper-parameters for baseline models evaluated on EgoTaskQA.}
\label{tab:hyper_parameter}
\resizebox{\linewidth}{!}{
    \begin{tabular}{cccccc}
    \toprule
    Model & Visual Input & Textual Input & Batch Size & Learning Rate & Training Iterations \\
    \midrule
     BERT & None & BERT embedding & 32 & $5 \times 10^{-5}$ & 25 epochs \\
     VisualBERT & ResNet & BERT embedding & 32 & $5 \times 10^{-5}$ & 40 epochs \\
     HGA & VGG+C3D & Glove embedding & 64 & $1 \times 10^{-4}$ &100 epochs \\
     HME & VGG+C3D & Glove embedding & 32 & $1 \times 10^{-3}$ & 25 epochs \\
     PSAC & ResNet & Glove embedding & 32 & $1 \times 10^{-3}$ & 50 epochs \\
     HCRN & ResNet + ResNext & Glove embedding & 32 & $1 \times 10^{-4}$ & 50 epochs \\
     HCRN w/o vision & None & Glove embedding & 32 & $1 \times 10^{-4}$ & 40 epochs \\
     ClipBERT & Grid Feature ResNet~\cite{jiang2020defense} & ClipBERT pretrained embedding & 24 & $2\times 10^{-5}$ & 25 epochs\\
    \bottomrule
    \end{tabular}
}
\end{table}

\section{Data Documentation}
We follow the datasheet proposed in~\cite{gebru2021datasheets} for documenting our EgoTaskQA benchmark:
\begin{enumerate}
    \item Motivation
    \begin{enumerate}
    \item For what purpose was the dataset created? \\
    This dataset was created to study goal-oriented task understanding in egocentric videos. Previous works lacks the data task-related object state and relationship annotations, as well as a good evaluation metric for such information.
    \item Who created the dataset and on behalf of which entity? \\
    This dataset was created by Baoxiong Jia, Ting Lei, Song-Chun Zhu and Siyuan Huang. At the time of creation, Baoxiong was a Ph.D. student at the \ac{ucla}, Ting was an undergraduate student at \ac{pku}, Siyuan was a research scientist at \ac{bigai}, and Song-Chun was a professor at \ac{ucla}, \ac{pku}, TsingHua University and \ac{bigai}.
    \item Who funded the creation of the dataset? \\
    The creation of this dataset was funded by \ac{bigai}.
    \item Any other Comments? \\
        A: None.
    \end{enumerate}
    \item Composition
    \begin{enumerate}
        \item What do the instances that comprise the dataset represent? \\
        For video data, each instance is a video clip provided in previous work LEMMA~\cite{jia2020lemma}. These videos record daily indoor activities. For question-answer pairs, each instance is consist of the question text, corresponding video interval, question scope, question type, targeting answer semantic and the program.
        \item How many instances are there in total? \\
        We crop videos in LEMMA into 2K video intervals for question answering. There are 40K question-answer pairs in total.
        \item Does the dataset contain all possible instances or is it a sample (not necessarily random) of instances from a larger set? \\
        We filter videos from the LEMMA dataset by removing videos with erroneous action annotations.
        \item What data does each instance consist of? \\
        See 2.(a).
        \item Is there a label or target associated with each instance? \\
        See 2.(a)
        \item Is any information missing from individual instances? \\
        No.
        \item Are relationships between individual instances made explicit? \\
        Video clips are related in the tasks performed in each videos as well as the performers. Question-answer pairs are related according to their metadata.
        \item Are there recommended data splits? \\
        For question answering, we provide two data splits \textit{normal} and \textit{indirect}. Refer to~\cref{sec:qa:generation} for more details.
        \item Are there any errors, sources of noise, or redundancies in the dataset? \\
        There are almost certainly some errors in video anntotations and question-answer pairs. We did our best to minimize these, but some certainly remain.
        \item Is the dataset self-contained, or does it link to or otherwise rely on external resources (e.g., websites, tweets, other datasets)? \\
        The dataset is self-contained.
        \item Does the dataset contain data that might be considered confidential (e.g., data that is protected by legal privilege or by doctor-patient confidentiality, data that includes the content of individuals' non-public communications)? \\
        No.
        \item Does the dataset contain data that, if viewed directly, might be offensive, insulting, threatening, or might otherwise cause anxiety? \\
        No.
        \item Does the dataset relate to people? \\
        Yes, all videos are recordings on human activities and all questions are related to these activities.
        \item Does the dataset identify any subpopulations (e.g., by age, gender)? \\
        No.
        \item Is it possible to identify individuals (i.e., one or more natural persons), either directly or indirectly (i.e., in combination with other data) from the dataset? \\
        Yes, we can recognize the actors in the original LEMMA recordings.
        \item Does the dataset contain data that might be considered sensitive in any way (e.g., data that reveals racial or ethnic origins, sexual orientations, religious beliefs, political opinions or union memberships, or locations; financial or health data; biometric or genetic data; forms of government identification, such as social security numbers; criminal history)?  \\
        No.
        \item Any other comments? \\
        None.
    \end{enumerate}
    \item Collection Process
    \begin{enumerate}
        \item How was the data associated with each instance acquired? \\
        We use the videos in LEMMA and generate question-answer pairs programmatically.
        \item What mechanisms or procedures were used to collect the data (e.g., hardware apparatus or sensor, manual human curation, software program, software API)? \\
        We use \ac{amt} to augment the original annotations in LEMMA.
        \item If the dataset is a sample from a larger set, what was the sampling strategy (e.g., deterministic, probabilistic with specific sampling probabilities)? \\
        See 2.(c).
        \item Who was involved in the data collection process (e.g., students, crowdworkers, contractors) and how were they compensated (e.g., how much were crowdworkers paid)? \\
        For video annotations, workers are paid at a rate of 1\$ per action annotation.
        \item Over what timeframe was the data collected? \\
        The videos were recorded by LEMMA and the question answer pairs were generated in summer 2022.
        \item Were any ethical review processes conducted (e.g., by an institutional review board)? \\
        No review processes were conducted with respect to the collection and annotation of this data.
        \item Does the dataset relate to people? \\
        Yes, see 2.(m).
        \item Did you collect the data from the individuals in question directly, or obtain it via third parties or other sources (e.g., websites)? \\
        We build websites for \ac{amt} workers to annotate the videos.
        \item Were the individuals in question notified about the data collection? \\
        Yes, we instruct the \ac{amt} workers to annotate all time-varying objects in the video intervals, as well as all multi-agent relationships on visibility and awareness.
        \item Did the individuals in question consent to the collection and use of their data? \\
        Yes, they were paid for these video annotations.
        \item If consent was obtained, were the consenting individuals provided with a mechanism to revoke their consent in the future or for certain uses? \\
        Yes, this is guaranteed by \ac{amt}.
        \item Has an analysis of the potential impact of the dataset and its use on data subjects (e.g., a data protection impact analysis) been conducted? \\
        No, all annotations are on objective world states with no subjective opinion or arguments involved.
        \item Any other comments? \\
        None.
    \end{enumerate}
    \item Preprocessing, Cleaning and Labeling
    \begin{enumerate}
        \item Was any preprocessing/cleaning/labeling of the data done (e.g., discretization or bucketing, tokenization, part-of-speech tagging, SIFT feature extraction, removal of instances, processing of missing values)? \\
        No, we annotated directly on the videos. For question-answer pair generation, we operate directly on the annotated videos.
        \item Was the "raw" data saved in addition to the preprocessed/cleaned/labeled data (e.g., to support unanticipated future uses)? \\
        Yes, we provide the raw data on our website.
        \item Is the software used to preprocess/clean/label the instances available? \\
        For video annotation, we adopt templates from \ac{amt}. We provide all other codes on our website.
        \item Any other comments? \\
        None.
    \end{enumerate}
    \item Uses
    \begin{enumerate}
        \item Has the dataset been used for any tasks already? \\
        No, the dataset is newly proposed by us.
        \item Is there a repository that links to any or all papers or systems that use the dataset? \\
        Yes, we provide the link to all related information on our website.
        \item What (other) tasks could the dataset be used for? \\
        The annotated videos could also be used for world model learning. The generated question-answer pairs could also be used for evaluating models' compositional reasoning capabilities.
        \item Is there anything about the composition of the dataset or the way it was collected and preprocessed/cleaned/labeled that might impact future uses? \\
        We propose to annotate the before/after status of each object given a video. We believe this could serve as a general protocol for annotating changing world states.
        \item Are there tasks for which the dataset should not be used? \\
        The usage of this dataset should be limited to the scope of activity or task understanding with its various downstream tasks (\eg action recognition, anticipation, state/relationship recognition and question answering).
        \item Any other comments? \\
        None.
    \end{enumerate}
    \item Distribution
    \begin{enumerate}
        \item Will the dataset be distributed to third parties outside of the entity (e.g., company, institution, organization) on behalf of which the dataset was created? \\
        Yes, the dataset will be made publicly available.
        \item How will the dataset will be distributed (e.g., tarball on website, API, GitHub)? \\
        The dataset could be accessed on our website.
        \item When will the dataset be distributed? \\
        The dataset will be released to the public upon acceptance of this paper. We provide private links for the review process.
        \item Will the dataset be distributed under a copyright or other intellectual property (IP) license, and/or under applicable terms of use (ToU)? \\
        We release our benchmark under CC BY-NC-SA\footnote{\url{https://paperswithcode.com/datasets/license}} license.
        \item Have any third parties imposed IP-based or other restrictions on the data associated with the instances? \\
        No.
        \item Do any export controls or other regulatory restrictions apply to the dataset or to individual instances? \\
        No.
        \item Any other comments? \\
        None.
    \end{enumerate}
    \item Maintenance
    \begin{enumerate}
        \item Who is supporting/hosting/maintaining the dataset? \\
        Baoxiong Jia is maintaining.
        \item How can the owner/curator/manager of the dataset be contacted (e.g., email address)? \\
        E-mail addresses are at the top of the paper.
        \item Is there an erratum? \\
        Currently, no. As errors are encountered, future versions of the dataset may be released and updated on our website.
        \item Will the dataset be updated (e.g., to correct labeling errors, add new instances, delete instances')? \\
        Yes, see 7.(c).
        \item If the dataset relates to people, are there applicable limits on the retention of the data associated with the instances (e.g., were individuals in question told that their data would be retained for a fixed period of time and then deleted)? \\
        No.
        \item Will older versions of the dataset continue to be supported/hosted/maintained? \\
        Yes, older versions of the benchmark will be maintained on our website.
        \item If others want to extend/augment/build on/contribute to the dataset, is there a mechanism for them to do so? \\
        Yes, errors may be submitted to us through email.
        \item Any other comments? \\
        None.
    \end{enumerate}
\end{enumerate}

\end{document}